\def\paperlanguage{}
\pdfoutput=1 % for arxiv

\documentclass[letterpaper, 10 pt, conference]{ieeeconf}  % Comment this line out if you need a4paper

\usepackage{bm}
\usepackage{cite}
\include{preamble}

\newcommand{\ctext}[1]{\raise0.2ex\hbox{\textcircled{\scriptsize{#1}}}}
\definecolor{bvqa}{rgb}{0.7,0.7,0}
\definecolor{mvqa}{rgb}{0.7,0,0}
\definecolor{itr}{rgb}{0.7,0,0.7}
\definecolor{vg}{rgb}{0,0,0.7}
\definecolor{dic}{rgb}{0,0.7,0.7}

\IEEEoverridecommandlockouts                              % This command is only needed if
\title{\LARGE \textbf
  {
    \switchlanguage%
    {%
      Robotic Applications of Pre-Trained Vision-Language Models\\to Various Recognition Behaviors
    }%
    {%
      事前学習済み視覚-言語モデルの認識行動に向けたロボティクスにおける応用
    }%
  }
}

\author{Kento Kawaharazuka$^{1}$, Yoshiki Obinata$^{1}$, Naoaki Kanazawa$^{1}$, Kei Okada$^{1}$, and Masayuki Inaba$^{1}$% <-this % stops a space
  \thanks{$^{1}$ The authors are with the Department of Mechano-Informatics, Graduate School of Information Science and Technology, The University of Tokyo, 7-3-1 Hongo, Bunkyo-ku, Tokyo, 113-8656, Japan.
    {\texttt\small [kawaharazuka, obinata, kanazawa, k-okada, inaba]@jsk.t.u-tokyo.ac.jp}
  }
}
\begin{document}

\maketitle
\thispagestyle{empty}
\pagestyle{empty}

%%%%%%%%%%%%%%%%%%%%%%%%%%%%%%%%%%%%%%%%%%%%%%%%%%%%%%%%%%%%%%%%%%%%%%%%%%%%%%%%
\begin{abstract}
  \switchlanguage%
  {%
    A number of models that learn the relations between vision and language from large datasets have been released.
    These models perform a variety of tasks, such as answering questions about images, retrieving sentences that best correspond to images, and finding regions in images that correspond to phrases.
    Although there are some examples, the connection between these pre-trained vision-language models and robotics is still weak.
    If they are directly connected to robot motions, they lose their versatility due to the embodiment of the robot and the difficulty of data collection, and become inapplicable to a wide range of bodies and situations.
    Therefore, in this study, we categorize and summarize the methods to utilize the pre-trained vision-language models flexibly and easily in a way that the robot can understand, without directly connecting them to robot motions.
    We discuss how to use these models for robot motion selection and motion planning without re-training the models.
    We consider five types of methods to extract information understandable for robots, and show the examples of state recognition, object recognition, affordance recognition, relation recognition, and anomaly detection based on the combination of these five methods.
    We expect that this study will add flexibility and ease-of-use, as well as new applications, to the recognition behavior of existing robots.
  }%
  {%
    近年, 視覚と言語の対応を大規模なデータセットから学習させたモデルが次々に公開されている.
    これらは, 画像に対する質問に答えるもの, 画像に最も良く対応する文章を検索するもの, 質問に対応する領域を画像中から見つけ出すもの等様々である.
    しかし, いくかの例が出始めてはいるものの, これら事前学習済みの視覚-言語モデルとロボットととの結びつきは未だ弱い.
    ロボットの動作に直接結びつけた場合, データ収集の困難さとロボットの身体性から汎用性が失われ, 多様な身体や状況に適応できないものとなってしまう.
    そこで本研究では, この事前学習済み視覚-言語モデルを直接ロボット動作とは結び付けず, ロボットが理解可能な形で柔軟かつ簡易に応用する方法について分類し, まとめることを行う.
    モデルを一切再学習せず, どうロボットの動作選択や動作計画に利用するかについて考察する.
    ロボットが理解可能な5種類の情報取得方法を考え, これらの組み合わせにより状態認識や物体認識, アフォーダンス認識や異常検知等を行い, その知見を示す.
    本研究が従来のロボットの認識行動に柔軟性と容易性を与えることを期待する.
  }%
\end{abstract}

\section{INTRODUCTION}\label{sec:introduction}
\switchlanguage%
{%
  With the advancement of neural network, various deep learning methods have been developed.
  The tasks include image recognition \cite{krizhevsky2012imagenet}, image generation \cite{goodfellow2014gan}, etc.
  Modalities such as vision and language are well suited for deep learning due to the ease of large-scale data collection, and they have made great progress mainly in the fields of computer vision and natural language processing.
  In addition, the development of vision-language models that combine these two aspects has been gaining popularity in recent years.
  \cite{antol2015vqa} has set forth a problem called Visual Question Answering (VQA) and released its dataset.
  \cite{radford2021clip} has successfully solved a zero-shot image classification problem by training the relations between images and phrases through contrastive learning.
  \cite{wang2022ofa} has successfully solved various problems such as Visual Question Answering, Image Captioning, and Visual Grounding with a single model.
  % However, it is regrettable that most of these models stay inside the computer, and that only accuracy is the competing factor.
  Several methods to learn the relations among various modalities is being applied to robot behavior.
  \cite{das2018eqa} has set forth a problem called Embodied Question Answering, in which the robot performs path planning to search for the answer to a question in a 3D simulation space.
  On the actual robot, \cite{hatori2018picking} has successfully achieved a picking task from object recognition and ambiguous verbal instructions.
  \cite{shridhar2021cliport} has enabled a variety of pick-and-place tasks using CLIP \cite{radford2021clip}.
  However, the data collection suddenly becomes difficult once the robot motion is used as data, and the number of data that can be contained in the dataset drops by a large order of magnitude.
  In most cases, the models trained using robots work only in the environment and robot body where the data is collected, resulting in a significant loss of versatility and adaptability.
  Note that there are efforts to solve this problem by collecting a large amount of data on a large number of embodiments \cite{google2023rtx}.
}%
{%
  深層学習の発展と伴に様々な学習手法が開発されてきた.
  特に, 画像認識\cite{krizhevsky2012imagenet}や画像生成\cite{goodfellow2014gan}等, そのタスクは様々である.
  視覚や言語といったモーダルは大規模なデータ収集が容易であるため, 深層学習との相性が良く, コンピュータビジョンや自然言語処理の分野を中心として大きく発展している.
  また, これら視覚や言語単体における学習だけでなく, 近年はこれらを結びつけた, 視覚-言語モデルの開発が盛んになりつつある\cite{li2022largemodels}.
  \cite{antol2015vqa}は画像に対してVisual Question Answering (VQA)という問題設定を打ち出し, データセット構築とそのベースライン手法の提案を行った.
  \cite{radford2021clip}は画像と言語の対応を対照学習により獲得し, zero-shotで画像分類問題を解くことに成功している.
  \cite{wang2022ofa}はVisual Question AnsweringやImage Captioning, Visual Grounding等の多様な問題を単一のモデルで解くことに成功している.
  しかし, これらのモデルは大抵コンピュータの中で完結し, その精度等を競うものになっている点が惜しい.
  一方で, この視覚や言語の多様なモダリティの関係性を学習する手法は, ロボットの動作にも結びつきつつある.
  \cite{das2018eqa}はEmbodied Question Answeringという問題設定を打ち出し, 3Dシミュレーション空間で, ロボットが質問の答えを探索する経路計画を行う.
  \cite{hatori2018picking}は実機において, 物体認識と曖昧な言語指示からロボットのピッキング動作を実現することに成功している.
  \cite{shridhar2021cliport}は言語指示をCLIP \cite{radford2021clip}により変換し多様なpick and place動作を可能にしている.
  しかし, 視覚や言語とは異なり, ロボットの動作をデータとして用いた途端に, そのデータ収集は突如として難しいものとなり, データセットに含まれるデータ数は大きく桁が落ちる.
  ロボットで学習させたモデルは, そのデータを取った環境や, そのロボットの身体でしか動かない場合がほとんどであり, 大きく汎用性や適応性を失う結果となってしまう.
  なお, それを大量の身体における大量のデータ収集により解決する取り組みも存在する\cite{google2023rtx}.
}%

\begin{figure}[t]
  \centering
  \includegraphics[width=0.8\columnwidth]{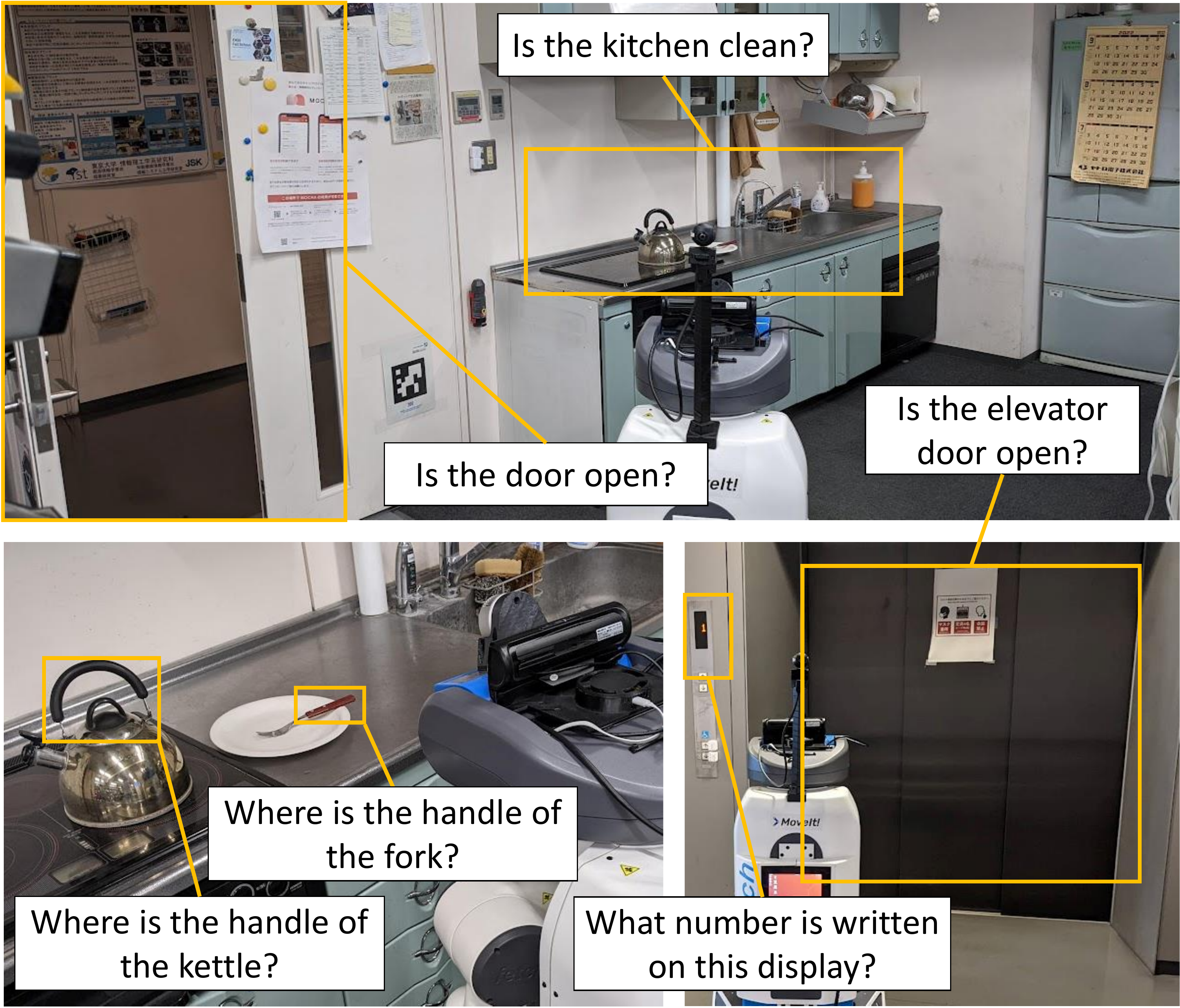}
  \vspace{-1.0ex}
  \caption{The concept of this study. The mobile robot recognizes if the door is open, if the kitchen is clean, where the handles of the kettle and fork are, what number is shown on the display, and if the elevator door is open.}
  \label{figure:concept}
  \vspace{-3.0ex}
\end{figure}

\switchlanguage%
{%
  From these points of view, we do not relate robot motions directly to vision and language in this study, but apply \textbf{Pre-Trained Vision-Language Models} (\textbf{PTVLMs}) in a form that can be understood by robots.
  The form of information understandable for robots is a continuous value with a few dimensions or a discrete value with a small number of choices.
  Depending on its usage, \textbf{PTVLMs} can significantly improve the recognition ability of existing robots, and lead to a variety of new recognition behaviors.
  We will categorize the extraction methods of information understandable for robots using \textbf{PTVLMs}, and experiment with various tasks that can be accomplished by combining the methods.
  Our policy is to take full advantage of the high versatility and adaptability of \textbf{PTVLMs} provided by the large datasets, so we do not perform any re-training using our own datasets that would reduce the versatility and adaptability.

  We will first describe some possible applications of this study.
  For example, whether the door is open or closed has been previously recognized by the presence or absence of point cloud, but by using \textbf{PTVLMs}, its recognition is possible by simply asking the question ``Is the door open?'' for the current image.
  The door handle has been previously recognized using template matching or a trained model with a dataset created by human annotation, but with \textbf{PTVLMs}, the location of a door handle can be recognized by simply asking ``Where is the door handle?''.
  Also, anomaly detection has been previously performed by learning images in the normal state and calculating the rate of reconstruction \cite{park2018anomaly}, but with \textbf{PTVLMs}, it is possible to detect anomalies with verbal explanations by comparing the descriptions of the normal and current states.
  Although a very simple idea, we believe that this method of applying \textbf{PTVLMs} can easily and flexibly increase the recognition ability of robots.

  % This study is organized as follows.
  % First, in \secref{sec:proposed}, we describe pre-trained vision-language models and its concrete usage.
  % Then, we categorize the five methods of extracting information in a form usable by robots, and summarize the various recognition behaviors that are possible by combining these methods.
  % In \secref{sec:experiment}, we show examples of recognition in actual everyday environments, and verify the high potential and effectiveness of our concept.
  % In \secref{sec:discussion}, we discuss the experimental results and conclude in \secref{sec:conclusion}.
}%
{%
  これらの観点から, 本研究ではロボットの動作と視覚や言語を直接関連付けせず, 大規模データセットにより事前学習済みの視覚-言語モデル(Pre-Trained Vision-Language Models, \textbf{PTVLMs})をロボットが理解可能な形で応用する.
  ここでいう理解可能な形式とは, 少次元の連続値または選択肢が少数に定まった離散値である.
  \textbf{PTVLMs}は使い方次第で既存のロボットの認識能力を柔軟かつ容易に大幅に向上させ, 多様な新しい認識行動を実現することに繋がる.
  ロボットが理解可能な\textbf{PTVLMs}の応用方法を分類し, その組み合わせにより可能になる多様なタスクについて紹介, 実験を行う.
  大規模データセットにより得られる高い汎用性と適応性を最大限に利用する方針であり, これらを下げる独自データセットを用いた一切の再学習は行わない.

  先に本研究によって可能となる応用事例について述べておく.
  例えば, これまでドアの開閉はポイントクラウドの有無により判断していたが, \textbf{PTVLMs}を用いて, ``ドアは空いていますか？''と質問するだけで, その開閉を認識できる.
  これまでコップやドアの取手はテンプレートマッチングや人間のアノテーションにより作成したデータセットを使い認識していたが, \textbf{PTVLMs}を用いて, ``ドアの取手はどこですか？''と質問するだけで, その場所を認識できる.
  これまで異常検知は正常状態において画像を集めて学習し, その画像の復元度により行っていたが\cite{park2018anomaly}, \textbf{PTVLMs}を用いることで, 正常状態と現在の状況説明を比較することで, 言語による説明つきの異常検知が可能になる.
  非常に単純なアイデアではあるが, この\textbf{PTVLMs}の応用方法はロボットの認識能力を容易かつ柔軟に, 大幅に引き上げると考えている.

  % 本研究の構成は以下である.
  % まず, \secref{sec:proposed}では, 事前学習済み視覚-言語モデルとその具体的な利用方法について述べる.
  % その後, ロボットに利用可能な形で情報を取得する方法を5つに分類し, これらを組み合わせることで可能となる様々な認識行動についてまとめる.
  % \secref{sec:experiment}では, 実際の日常環境における認識例を示し, 本研究のコンセプトの高いポテンシャルと有効性を検証する.
  % \secref{sec:discussion}では実験結果について考察し, \secref{sec:conclusion}で結論を述べる.
}%

\begin{figure}[t]
  \centering
  \includegraphics[width=1.0\columnwidth]{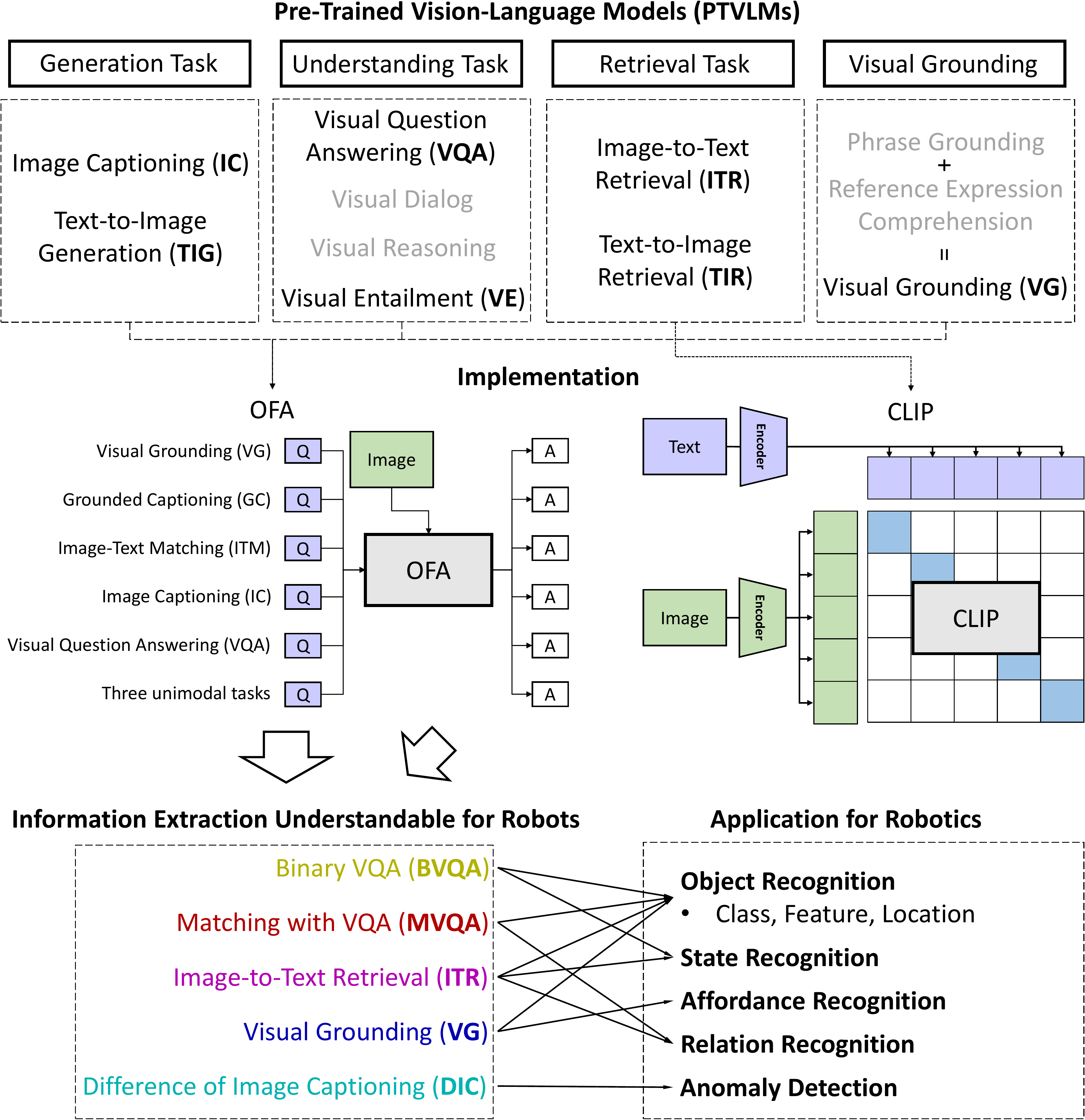}
  \vspace{-3.0ex}
  \caption{The overview of tasks using \textbf{Pre-Trained Vision-Language Models} (\textbf{PTVLMs}), implementations of \textbf{PTVLMs}, information extraction understandable for robots using \textbf{PTVLMs}, and robotic applications of the information extractions.}
  \label{figure:proposed}
  \vspace{-3.0ex}
\end{figure}

\section{Robotic Applications of Pre-Trained Vision-Language Models} \label{sec:proposed}
\subsection{Overview of Pre-Trained Vision-Language Models} \label{subsec:model-overview}
\switchlanguage%
{%

  The overall structure of this study is shown in \figref{figure:proposed}.
  There are various tasks in the vision-language model, and \cite{li2022largemodels} has classified them into four categories: generation task, understanding task, retrieval task, and grounding task.
  The generation task includes Image Captioning (IC), which generates captions of images, and Text-to-Image Generation (TIG), which generates images from texts.
  The understanding task includes Visual Question Answering (VQA), which answers questions about images, Visual Dialog, which answers questions based on images and dialog history, Visual Reasoning, which requires justification to the answer of VQA, and Visual Entailment (VE), which verifies the semantic validity of image-text pairs.
  The retrieval task includes Image-to-Text Retrieval (ITR) and Text-to-Image Retrieval (TIR), which retrieve the relationship between image and text from predefined choices.
  The grounding task includes Phrase Grounding and Reference Expression Comprehension, which extract the bounding box of the corresponding location in the image from the text.

  Considering the simplicity of use for robots, we only handle tasks that use direct vision-language relations.
  In other words, we do not deal with Visual Dialog which includes dialog history and Visual Reasoning which requires reasoning.
  To summarize, the tasks we handle in this study are IC, TIG, VQA, VE, ITR, TIR, and VG (since Phrase Grounding and Reference Expression Comprehension are almost the same, we unify them as Visual Grounding, VG).
  In this study, we treat OFA \cite{wang2022ofa} and CLIP \cite{radford2021clip} as \textbf{PTVLMs} that cover these tasks.
  OFA is a method that unifies IC, TIG, VQA, VE, and VG with a single model.
  CLIP is a model that can acquire the relationship between images and texts through contrastive learning, and is capable of ITR and TIR.
  Note that the model itself is not limited to OFA or CLIP, as long as it can perform these tasks.
}%
{%
  本研究の全体像を\figref{figure:proposed}に示す.
  Vision-Language Modelのタスクは様々であるが, \cite{li2022largemodels}ではこれをGeneration Task, Understanding Task, Retrieval Task, Grounding Taskの4つに分類している.
  Generation Taskには, 画像のキャプションを生成するImage Captioning (IC)と, 言語から画像を生成するText-to-Image Generation (TIG)が含まれる.
  Understanding Taskには, 画像に対する質問に回答するVisual Question Answering (VQA), 画像と対話履歴から質問に回答するVisual Dialog, VQAに加えてその理由を回答する必要のあるVisual Reasoning, 画像と言語のペアに対してその意味的な妥当性を検証するVisual Entailment (VE)が含まれる.
  Retrieval Taskには, 画像と言語の対応を選択肢から検索するImage-to-Text Retrieval (ITR)とText-to-Image Retrieval (TIR)が含まれる.
  Grounding Taskには, 言語から画像中の対応する箇所のバウンディングボックスを抜き出すPhrase GroundingとReference Expression Comprehensionが含まれる.

  この中でも, ロボットへの簡易な利用を考え, 本研究では視覚-言語間の直接的な関係性を用いるタスクに限定する.
  つまり, 対話履歴を含むVisual Dialogや, 理由を回答するVisual Reasoningについては扱わない.
  整理すると, 本研究で扱うタスクはIC, TIG, VQA, VE, ITR, TIR, VG (Phrase GroundingとReference Expression Comprehensionはほとんど同一であるため, Visual Grounding, VGとして統一した)である.
  これらを網羅する事前学習済み視覚-言語モデルとして, 本研究ではOFA \cite{wang2022ofa}とCLIP \cite{radford2021clip}を扱う.
  OFAはIC, TIG, VQA, VE, VGを単一モデルによって統一的に行う手法である.
  また, CLIPは画像と言語の関係を対照学習により獲得でき, ITRとTIRが可能なモデルである.
  なお, これらタスクが可能なモデルであれば, そのモデル自体はOFAやCLIPに限定されない.
}%

\subsection{Information Extraction from Pre-Trained Vision-Language Models for Robotics} \label{subsec:information-extraction}
\switchlanguage%
{%
  Discrete values with multiple choices (e.g. language) are difficult to handle in a robot, which operates by human programming.
  Also, multi-dimensional continuous values (e.g. vision) are difficult to incorporate directly into a program, and some kind of information extraction is necessary.
  In other words, it is necessary to extract information as discrete values in a predefined small number of choices, or as continuous values in a small number of dimensions.
  In this study, we classify the methods to extract information from \textbf{PTVLMs} in a way that robots can understand into the following five categories.
  \begin{itemize}
    \item \textbf{\textcolor{bvqa}{Binary VQA} (\textcolor{bvqa}{BVQA})}: binary information is extracted by asking questions such as ``Is -?'' and ``Do -?'' in VQA.
    \item \textbf{\textcolor{mvqa}{Matching with VQA} (\textcolor{mvqa}{MVQA})}: discrete values in a small number of choices are obtained by asking questions such as ``What kind of -?'' and ``How much -?'' in VQA and by matching the answers to the phrases prepared in advance.
    \item \textbf{\textcolor{itr}{Image-to-Text Retrieval} (\textcolor{itr}{ITR})}: discrete values in a small number of choices are obtained by searching for the phrase that best matches the image among the predefined choices.
    \item \textbf{\textcolor{vg}{Visual Grounding} (\textcolor{vg}{VG})}: continuous values in a small number of dimensions are obtained by selecting the bounding box in the image that best matches the phrase.
    \item \textbf{\textcolor{dic}{Difference of Image Captioning} (\textcolor{dic}{DIC})}: binary information is obtained by quantifying the difference between the descriptions output by IC.
  \end{itemize}
  It is difficult to extract information directly from images in a form understandable by robots, so it is difficult to use Text-to-Image Generation (TIG) and Text-to-Image Retrieval (TIR).
  If the output of VQA is a sentence, it is not possible to extract information directly in a form understandable by robots, so we ask ``Is -?'' or ``Do -?'' questions that return a binary answer of Yes or No (\textcolor{bvqa}{BVQA}), ask ``What kind of -?'' or ``How much -?'' questions that output simple and short answers that match predefined choices (\textcolor{mvqa}{MVQA}), or calculate the difference in descriptions output by Image Captioning using Sentence-BERT \cite{reimers2019sentencebert} (\textcolor{dic}{DIC}).
  \textcolor{itr}{ITR} and \textcolor{vg}{VG} can directly output the answers in a form understandable by robots.
  Visual Entailment (VE) is considered to be included in \textcolor{bvqa}{BVQA}.

  The details of the models to be used for the five types of information extraction are described below.
  % For \textcolor{bvqa}{BVQA}, \textcolor{mvqa}{MVQA}, and \textcolor{dic}{DIC}, we use the fine-tuned model with COCO dataset \cite{lin2014coco} based on the pre-trained model OFA$_{large}$.
  % For \textcolor{vg}{VG}, we use the fine-tuned model with RefCOCOg dataset \cite{mao2014refcocog} based on the pre-trained model OFA$_{large}$.
  % OFA is pre-trained with five multimodal tasks and three unimodal tasks.
  % OFA is pre-trained with five multimodal tasks: VQA, IC, \textcolor{vg}{VG}, ITM (Image-Text Matching - answering Yes or No questions), GC (Grounded Captioning - the reverse of \textcolor{vg}{VG}, performing IC for a bounding box of the image), and with three unimodal tasks.
  % We use a single model to extract information with \textcolor{bvqa}{BVQA}, \textcolor{mvqa}{MVQA}, and \textcolor{dic}{DIC} for the sake of simplicity.
  % Although fine-tuned models for such as ITM and VQA can be constructed, since all of them showed high performance, we use a single model to extract information with \textcolor{bvqa}{BVQA}, \textcolor{mvqa}{MVQA}, and \textcolor{dic}{DIC} for the sake of simplicity.
}%
{%
  ロボットという, 人間によりプログラムされることで動く機械において, 言語のような離散値の集合はそのままでは非常に扱いにくい.
  また, 視覚のような多次元の連続値も, これを直接プログラムに組み込むことは難しく, 何らかの情報抽出が必要である.
  つまり, 予め定まった少数選択肢における離散値, または, 少次元の連続値として情報取得する必要がある.
  そこで本研究では, ロボットが理解可能な形で\textbf{PTVLMs}から情報抽出を行う方法について以下のように5つに分類する.
  \begin{itemize}
    \item Binary VQA (\textcolor{bvqa}{BVQA}): VQAにおいて``Is -?''や``Do -?''といった質問を行い, ロボットが理解可能な二値で情報を取得する.
    \item Matching with VQA (\textcolor{mvqa}{MVQA}): VQAにおいて``What kind of -?''や``How much -?''といった質問を行い, 答えに対して予め用意したフレーズとマッチングすることで, 少数選択肢における離散値を得る.
    \item Image-to-Text Retrieval (\textcolor{itr}{ITR}): 画像に最もマッチするフレーズを選択肢の中から検索することで, 少数選択肢における離散値を得る.
    \item Visual Grounding (\textcolor{vg}{VG}): フレーズに最もマッチする画像中のバウンディングボックスを選ぶことで, 少次元の連続値を得る.
    \item Difference of Image Captioning (\textcolor{dic}{DIC}): 画像に対して得られた状況説明の差を数値化することで, 二値で情報を取得する.
  \end{itemize}
  画像からはロボットが理解可能な形で直接情報抽出することが難しいため, TIGやTIRを用いるのは難しい.
  VQAの出力が文章である場合は直接ロボットが理解可能な情報抽出ができないため, ``Is -?''や``Do -?''等のYesまたはNoの二値を返す質問をしたり, ``What kind of -?''や``How much -?''等の短いフレーズや一単語で答えが返ってくる質問をして選択肢とマッチングしたり, または``What -?''による質問やImage Captioningにおいて出力された文章についてその差分をSentence-BERT等で計算する, という選択肢が考えられる.
  また, \textcolor{itr}{ITR}や\textcolor{vg}{VG}は直接ロボットにとって理解可能な形で答えを利用することができる.
  なお, VEは\textcolor{bvqa}{BVQA}に含まれると考えられる.

  5つの情報抽出について詳細を以下に述べる.
  なお, \textcolor{bvqa}{BVQA}, \textcolor{mvqa}{MVQA}, \textcolor{dic}{DIC}はOFAの事前学習済みモデルOFA$_{large}$について, COCOデータセット\cite{lin2014coco}を用いたFine Tuningモデルを用いる.
  \textcolor{vg}{VG}については, OFAの事前学習済みモデルOFA$_{large}$について, RefCOCOgデータセット\cite{mao2014refcocog}を用いたFine Tuningモデルを用いる.
  OFAはVQA, IC, \textcolor{vg}{VG}, ITM (Image-Text Matching, Yes/Noで回答可能な質問への回答), GC (Grounded Captioning, \textcolor{vg}{VG}の逆でバウンディングボックス内のICを行う)の5つのタスクと3つの単一モーダルタスクにより事前学習されている.
  ITMやVQA等にFine-Tuningされたモデルも構成可能であるが, どれも高い性能を示したため, 簡易さから一つのモデルで\textcolor{bvqa}{BVQA}, \textcolor{mvqa}{MVQA}, \textcolor{dic}{DIC}の３つの情報抽出を行っている.
}%

\subsubsection{\textbf{\textcolor{bvqa}{Binary VQA}}}
\switchlanguage%
{%
  \textcolor{bvqa}{BVQA} is a task that asks questions $Q$ such as ``Is the door open?'' or ``Does the image describe a person running?'' for the current image $V$ to extract binary answers $A$ of Yes or No.
  We can express \textcolor{bvqa}{BVQA} as $\bm{f}_{BVQA}(V, Q)=A$ ($A\in\{Yes, No\}$).
  Binary values are easy to use with the robot programming of ``if-else''.
  Multiple $V$ with noise and multiple $Q$ with rephrased expressions of the same content are prepared, and $A$ is determined based on the ratio of Yes and No.
  In this study, five $V$ are prepared by RGBShift, which adds randomly selected values from a uniform distribution within the range of [-0.1, 0.1] to each RGB value.
  We prepare four types of $Q$ by changing the article to \{a, the, this, that\}, and integrate a total of 20 results which are the combinations of these $V$ and $Q$.
  Since OFA is trained by combining five multimodal tasks including IC and VQA, $A$ does not always become Yes or No.
  Therefore, if a value other than Yes or No is obtained, the answer is considered as Invalid.
}%
{%
  \textcolor{bvqa}{BVQA}は例えば, 画像$V$に対して``Is the door open?''や``Does the image describe a person running?''等の質問$P$をし, YesまたはNoの二値の結果を得るタスクである.
  二値の値はロボットの``if-else''との相性も良く使いやすい.
  ノイズを加えた複数の画像と, 同じ内容の表現を言い換えた複数の質問を用意し, YesとNoの割合から答えを決定する.
  本研究ではノイズの加え方として, RGBそれぞれの値に[-0.1, 0.1]の範囲内の一様分布からランダムに選ばれた値を足し込むRGBShiftにより5つの画像を用意する.
  質問については冠詞を\{a, the, this, that\}に変更することで4種類の質問を用意し, これらの掛け合わせである合計20の結果を総合する.
  また, VQAやIC, ITMなどを複合して学習したモデルであるため, YesまたはNoが必ず出力されるわけではない.
  そのため, YesまたはNoでない値が出た場合は答えを無効とする.
}%

\subsubsection{\textbf{\textcolor{mvqa}{Matching with VQA}}}
\switchlanguage%
{%
  \textcolor{mvqa}{MVQA} is a task to select the matching answer $A$ to questions $Q$ such as ``What object is included in this image?'', ``What color is the apple?'', and ``How big is this apple?'' from the predefined choices $C$.
  We can express \textcolor{mvqa}{MVQA} as $\bm{f}_{MVQA}(V, Q, C_{\{1, \cdots, N_{C}\}})=A$ ($A\in C_{\{1, \cdots, N_{C}\}}$, where $N_{C}$ expresses the number of predefined choices).
  The robot behavior is described for each $C$, and conditional branching is possible depending on the selected $C$.
  Similar to \textcolor{bvqa}{BVQA}, multiple $V$ with noise and multiple $Q$ about the same state are prepared, and $A$ is determined based on the percentage of matches.
  Since the $C$ are not always output as the answers, the answers are considered Invalid if they do not match.
}%
{%
  \textcolor{mvqa}{MVQA}は例えば, ``What object is included in this image?''や``What color is the apple?'', ``How big is this apple?''などの答えに対して, 合致する選択肢を選ぶタスクである.
  選択肢それぞれに対してロボットの行動を記述しておき, どの選択肢が選ばれるかに応じて条件分岐が可能である.
  \textcolor{bvqa}{BVQA}と同様に, ノイズを加えた複数の画像と同じ事象に関する複数の質問を用意し, 合致した割合から答えを決定する.
  予め記述した選択肢が必ず答えとして出力されるわけではないため, 合致しない場合は答えを無効とする.
}%

\subsubsection{\textbf{\textcolor{itr}{Image-to-Text Retrieval}}}
\switchlanguage%
{%
  \textcolor{itr}{ITR} calculates the degree of matching between the current image $V$ and each of the predefined phrase choices $C$, and selects the choice $A$ with the highest degree of matching.
  We can express \textcolor{itr}{ITR} as $\bm{f}_{ITR}(V, C_{\{1, \cdots, N_{C}\}})=A$ ($A\in C_{\{1, \cdots, N_{C}\}}$, where $N_{C}$ expresses the number of the predefined choices).
  Similarly to \textcolor{mvqa}{MVQA}, the actions are described for each $C$, and conditional branching is possible depending on the selected $C$.
  In the original CLIP to conduct \textcolor{itr}{ITR}, a sentence similar to the current sentence is selected from a large amount of textual datasets.
  However, in this study, a small number of $C$ are given in advance and the robot selects its answer among them, so that it is understandable for robots.
  Compared to \textcolor{mvqa}{MVQA}, it is likely to be able to select a more precise answer than \textcolor{mvqa}{MVQA}, since it can output the probability for each $C$.
}%
{%
  \textcolor{itr}{ITR}はCLIPを用いて行う.
  現在の画像と予め記述した選択肢の一致度をそれぞれ計算し, 一致度が最も高い選択肢を選ぶ.
  \textcolor{mvqa}{MVQA}と同様に, 選択肢それぞれに対して行動を記述しておき, どの選択肢が選ばれるかという条件分岐が可能である.
  なお, 本来のCLIPでは多量のテキストデータセットから現在の文に近い文を選択するが, 本研究ではロボットが理解可能という意味で, 予め少数の選択肢を与えその中で回答を選択する.
  \textcolor{mvqa}{MVQA}に比べ, それぞれの選択肢に関する確率を出すことができるため, より精度良く選択肢を選ぶことができる可能性が高い.
}%

\subsubsection{\textbf{\textcolor{vg}{Visual Grounding}}}
\switchlanguage%
{%
  \textcolor{vg}{VG} is a task to determine a target phrase $Q$ such as ``a refrigerator'' to ask the question ``Which region does the text `a refrigerator' describe?'', and to get the bounding box $A$ of the relevant part.
  We can express \textcolor{vg}{VG} as $\bm{f}_{VG}(V, Q)=A$ ($A$ expresses the coordinates of the bounding box).
  This method is compatible with various controls because it can obtain the location of an object in $V$ based on an arbitrary $Q$.
  Note that there are many methods to perform only \textcolor{vg}{VG}, such as ViLD \cite{gu2022vild} and LSeg \cite{li2022lseg}.
}%
{%
  \textcolor{vg}{VG}は例えば, ``a refrigerator''等のフレーズを決め, OFAに対して``Which region does the text `a refrigerator' describe?''という質問をすることで, 該当する箇所のバウンディングボックスを得ることができる.
  任意のフレーズの画像上の物体位置を取得できるため, 様々な制御と相性が良い.
  なお, ViLD \cite{gu2022vild}やLSeg \cite{li2022lseg}など, \textcolor{vg}{VG}のみを行う手法は多く存在する.
}%

\subsubsection{\textbf{\textcolor{dic}{Difference of Image Captioning}}}
\switchlanguage%
{%
  \textcolor{dic}{DIC} computes the difference $A$ of the situational descriptions obtained from the question $Q$ ``What does the image describe?'' for two images $V_{\{1, 2\}}$.
  We can express \textcolor{dic}{DIC} as $\bm{f}_{DIC}(V_{\{1, 2\}}, Q)=A$ ($-1\leq{A}\leq1$).
  The two sets of the situational descriptions obtained at a certain location are vectorized by Sentence-BERT \cite{reimers2019sentencebert}, etc.
  By calculating the cosine similarity $A$ between these vectors and cutting them by a threshold value, it is possible to obtain a binary result of whether or not a change has occurred.
  In addition, since the language is given, it is easy for humans to understand the change intuitively.
  % In this study, Doc2Vec is pre-trained with text8 dataset and with the dimension of the vector set to 40.
  Also, by using GPT-3 \cite{brown2020gpt3} and asking questions such as ``What is the difference between `text1' and `text2'?'' (where `text1' and `text2' refer to each situational description), it is possible to directly output what the difference between them is as sentences.
}%
{%
  \textcolor{dic}{DIC}は, OFAにおいて``What does the image describe?''の質問から得られた状況説明文の差分を計算する.
  それぞれの場所について得られた二回分の状況説明文章をSentence-BERT \cite{reimers2019sentencebert}等によりベクトル化する.
  このベクトル同士のコサイン類似度を計算し閾値で切ることで変化したかどうかを二値で得ることができる.
  また, 言語が与えられるため, 人間が直感的に変化を理解しやすい.
  % なお, 本研究ではDoc2Vecをtext8データセットを用いて, ベクトルの次元を40として事前に学習させた.
  GPT-3 \cite{brown2020gpt3}などを用いて, ``What is the difference between `text1' and `text2'?''等の質問をすることで(ここで, text1とtext2はそれぞれ状況説明文を指す), 何が違うのかを直接文章として出力することも可能である.
}%

\subsection{Robotic Applications of Information Extraction from Pre-Trained Vision-Language Models} \label{subsec:robotic-application}
\switchlanguage%
{%
  Using the five information extraction methods described so far, we categorize the specific tasks that can be performed.
  Here, we consider the following five tasks: object recognition, state recognition, affordance recognition, relation recognition, and anomaly detection.
}%
{%
  これまでに説明したロボットが理解可能な５つの情報抽出手段を用いて, 具体的に可能となるタスクについて分類しまとめる.
  ここでは物体認識, 状態認識, アフォーダンス認識, 関係認識, 異常検知の5つについて考える.
}%

\subsubsection{\underline{\textbf{Object Recognition}}}
\switchlanguage%
{%
  Among object recognition, we can mainly perform class recognition, feature recognition, and location recognition of objects.
  Class recognition can be achieved by making a binary judgment of whether a particular object exists or not using \textcolor{bvqa}{BVQA}, or by making a multiple-choice judgment using \textcolor{mvqa}{MVQA} or \textcolor{itr}{ITR}.
  This method can be used to determine whether or not an object is correctly grasped and to make a transition to the next action.

  Feature recognition can be achieved by a multiple-choice judgment using \textcolor{mvqa}{MVQA} or \textcolor{itr}{ITR}.
  $Q$ of ``How big is -?'' or ``What color is -?'' in \textcolor{mvqa}{MVQA} can extract information such as color, shape, and size of an object.
  Regarding \textcolor{itr}{ITR}, feature recognition is also possible by defining $C$ such as ``a \{large, medium, small\} egg''.

  Location recognition can be achieved by outputting the location of a specific object in the image using \textcolor{vg}{VG}.
  The obtained locations can be used for grasping control by masking the point cloud within the location, or for wheeled-base control by measuring the distance to the location.

  By combining these methods, stepwise refinement of object recognition is also possible.
  The presence of the target object is confirmed by \textcolor{bvqa}{BVQA}, \textcolor{mvqa}{MVQA} or \textcolor{itr}{ITR}, and its bounding box is cropped by \textcolor{vg}{VG}.
  After that, class recognition for another target object against the cropped image is performed, and these process are executed iteratively.
  This method is useful for object recognition in a refrigerator, object recognition on a desk at a distance, and so on.
}%
{%
  物体認識の中でも主に, 物体のクラス認識, 特徴認識, 位置認識が可能である.
  物体のクラス認識は, \textcolor{bvqa}{BVQA}による特定の物体が存在するかどうかの二値判断や, \textcolor{mvqa}{MVQA}や\textcolor{itr}{ITR}による複数選択肢判断により可能となる.
  正しい物体を掴めているかや, 次の行動への遷移に利用できる.

  物体の特徴認識は, \textcolor{mvqa}{MVQA}や\textcolor{itr}{ITR}による複数選択肢判断により可能となる.
  \textcolor{mvqa}{MVQA}における``How big -?''や``What color -?''と言った質問により, 物体の色や形, 大きさ等の情報抽出が可能である.
  \textcolor{itr}{ITR}についても, ``a \{large, middle, small\} egg''などの選択肢を定義することで, 特徴認識が可能である.

  物体の位置認識は, \textcolor{vg}{VG}による特定の物体に関する画像中の位置出力により可能となる.
  得られた位置からポイントクラウドをマスクし把持制御に用いることや, 該当する場所までの距離を測りロボット制御に利用可能である.

  また, これらを組み合わせることで詳細な物体認識も可能である.
  \textcolor{bvqa}{BVQA}, \textcolor{mvqa}{MVQA}, \textcolor{itr}{ITR}等により対象物体の存在を確認し, \textcolor{vg}{VG}によりそのバウンディングボックスを切り取る.
  その後, 切り取られた画像に対して対して別の対象物体に関するクラス認識を行う, というように順を追った詳細な物体認識が可能である.
  冷蔵庫中の物体認識や, 少し離れた机の上の物体認識等に有用である.
}%

\subsubsection{\textbf{\underline{State Recognition}}}
\switchlanguage%
{%
  State recognition mainly uses \textcolor{bvqa}{BVQA} or \textcolor{itr}{ITR} to check the binary state of the current environment, objects, etc.
  For example, in \textcolor{bvqa}{BVQA}, the state of a door can be recognized by asking ``Is this door open?''.
  Regarding \textcolor{itr}{ITR}, the same result can be obtained by preparing $C$ of ``an open door'' and ``a closed door''.
  Another benefit of this state recognition is that it can recognize ambiguous and qualitative states based on the knowledge obtained from a large-scale dataset.
  For example, it can make human-like judgments for $Q$ such as ``Is this kitchen clean?''.

  In addition, it can recognize a character state by asking $Q$ such as ``What is written on this bottle?''.
  Moreover, this state recognition is likely to make it possible to write conditional expressions such as ``if-else'' and ``assert'' in robot programming using the spoken language.
}%
{%
  状態認識は主に, 現在の環境や物体等について, \textcolor{bvqa}{BVQA}や\textcolor{itr}{ITR}を使いその状態を確認するものである.
  例えば\textcolor{bvqa}{BVQA}において, ``Is this door open?''と質問することで, ドアの開閉状態を得ることができる.
  これは\textcolor{itr}{ITR}において, ``an open door''と``a closed door''の2つの選択肢を用意することでも同様の結果を得ることができる.
  また, この状態認識は不明瞭で質的な状態を大規模なデータから得られた知見により認識できる点も魅力的である.
  例えば, ``Is this kitchen clean?''や, ``Is this onion roasted?''など, これまでの状態認識には無かった人間らしい判断を行うことができる.

  その他, ``What is written in this bottle?''という質問による特定部分の文字認識や, ``What number is shown in the elevator display?''という質問による数字の認識も可能である.
  また, 本状態認識によって, ロボットプログラミングにおけるif-elseやassertの条件式を, 言語によって記述することができるようになる可能性が高い.
}%

\subsubsection{\textbf{\underline{Affordance Recognition}}}
\switchlanguage%
{%
  Affordance refers to the role of each part of a tool or object; for example, the place to grasp or the effector that exerts the action.
  \textbf{PTVLMs} enable implicit affordance recognition, whereas in most cases, a dedicated network for affordance recognition has been set up so far \cite{myers2015affordance}.
  For example, by running \textcolor{vg}{VG} on $Q$ such as ``handle of the scissors'' or ``handle of the kettle'', it is possible to recognize the parts of tools and objects with meaning related to their operations.
  The recognized parts can be used for grasping, tilting the spout of the kettle toward the cup, and so on.
}%
{%
  Affordanceとは, 道具や物体のそれぞれの部位における意味, 例えば, 掴む場所や作用を及ぼす効果器等を現したものである.
  これまでAffordance認識専用のネットワークを組むことがほとんどであったが\cite{myers2015affordance}, \textbf{PTVLMs}により暗黙的にAffordance認識が可能となる.
  例えば, ``a handle of the scissors''や, ``a handle of the kettle''等のフレーズに対して\textcolor{vg}{VG}を実行することで, 意味を持った道具の部位を認識することができる.
  認識した箇所を掴んだり, やかんにおける水の出口をコップ方向に傾けたり等の利用が可能である.
}%

\begin{figure*}[t]
  \centering
  \includegraphics[width=2.0\columnwidth]{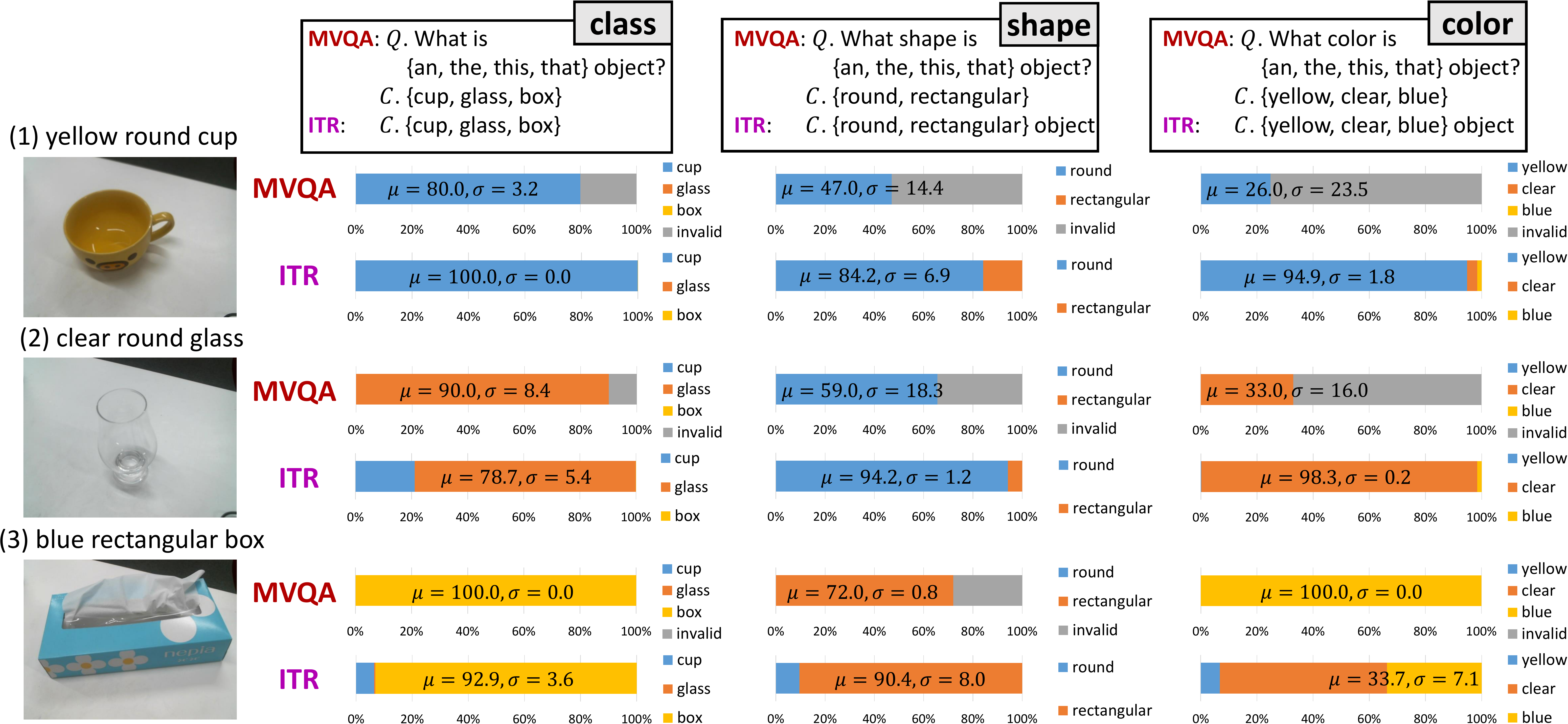}
  \vspace{-1.0ex}
  \caption{The results of the object class, shape, and color recognition using \textcolor{mvqa}{MVQA} or \textcolor{itr}{ITR} for three examples.}
  \label{figure:object-each}
  \vspace{-3.0ex}
\end{figure*}

\subsubsection{\textbf{\underline{Relation Recognition}}}
\switchlanguage%
{%
  \textcolor{mvqa}{MVQA} and \textcolor{itr}{ITR} allow us to recognize relationships between objects.
  In \textcolor{mvqa}{MVQA}, by asking $Q$ ``What is the relative relationship between the mouse and keyboard?'', we can extract the relationship in terms of position, such as ``next to'', ``in front of'', ``on'', and ``under''.
  By recognizing the relations based on which prepositions are included in the sentences, we can generate the next robot movements.
  This can also be used for segmentation of motions.
  By recognizing not only changes in relationships between objects but also relationships between the robot and objects, the robot can detect the changing point as an anomaly by using \textcolor{dic}{DIC}.
  % There is a high possibility that this method can be applied to human behavior segmentation and imitation learning.
}%
{%
  \textcolor{mvqa}{MVQA}や\textcolor{itr}{ITR}により, 物体同士の関係性を認識することができる.
  \textcolor{mvqa}{MVQA}において, ``What is the relative relationship between the mouse and keyboard?''という質問をすることで, それらの関係性である``next to''や``in front of'', ``on''や``under''等の関係性を抜き出すことができる.
  どの前置詞が文に含まれるかによって関係性を認識し, ロボットの動きを生成することができる.
  そしてこれは, 動作の分節化にも利用することが可能である.
  物体同士の関係性の変化だけでなく, ロボットと物体の間の関係性を認識することで, その変化点を\textcolor{dic}{DIC}と同様に異常度として検知する.
  % 模倣学習や人間の行動認識にも応用できる可能性が高い.

  % TODO: 検証
  % V+ Q -> A
  % V+ Q + M -> A
  % V+ M -> A
  % V+ P -> A
  % V_1, V_2 + Q -> A
}%

\subsubsection{\textbf{\underline{Anomaly Detection}}}
\switchlanguage%
{%
  \textcolor{dic}{DIC} can be used for anomaly detection.
  Until now, anomaly detection has usually been performed by collecting and learning images in normal conditions, and then using the degree of reconstruction of the images \cite{park2018anomaly}.
  Here, in the normal state, the robot takes pictures at a certain location and records them.
  Next, we generate captions for the current images taken at the same location and for the images taken in the normal state, and output the difference between them using \textcolor{dic}{DIC}.
  If the difference is large, it is assumed to be abnormal, and humans can understand the change linguistically from the difference between the captions.
}%
{%
  \textcolor{dic}{DIC}を用いることで, 異常検知を行うことができる.
  これまで異常検知は正常状態において画像を集めて学習し, その画像の復元度により行うことが多かったが\cite{park2018anomaly}, これを言語から行う.
  ロボットが正常な状態において決まった場所で写真を撮り記録しておく.
  次に, 現在状態において同一の場所について撮られた写真と正常時の写真のキャプションを生成し, \textcolor{dic}{DIC}によりその差を出力する.
  差が大きい場合は異常であるとして, キャプション間の差から人間が言語的にその変化を理解することができる.
}%

\section{Preliminary Experiments} \label{sec:experiment}
\switchlanguage%
{%
  We will experimentally describe specific use cases for the five recognition behaviors described in \secref{subsec:robotic-application}.
  The purpose of this study is to classify the usage of \textbf{PTVLMs} and its possible applications for easily and flexibly generating recognition behaviors of robots.
  The experiments are mainly qualitative as the comprehensiveness of the applications is most important.
  On the other hand, the reliability of each experimental application is ensured by adding noise to the images or changing the angle of view to compute the mean and variance.
}%
{%
  \secref{subsec:robotic-application}で述べた5つの認識行動について具体的な利用事例を実験的に述べる.
  本研究の趣旨は簡易で柔軟な認識行動生成に向けた\textbf{PTVLMs}の利用方法と可能となるロボットタスクの分類であり, 定性的な実験が主体である.
  一方で, それぞれの応用実験例については, 画像にノイズを加えたり画角を変えて, 平均と分散を出すことで, その信頼性を担保している.
  % TODO: なお, 本研究では本コンセプトの広い有効性を示すのみとし...
}%

\begin{figure}[t]
  \centering
  \includegraphics[width=0.95\columnwidth]{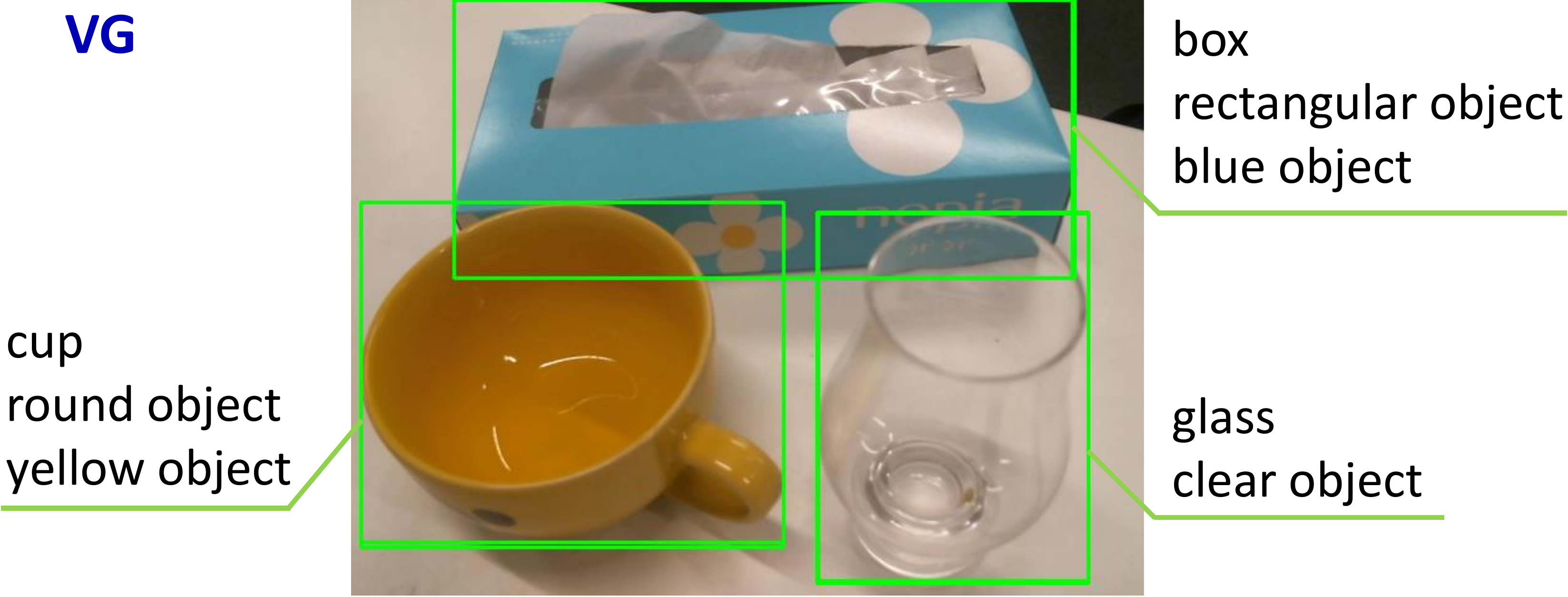}
  \vspace{-1.0ex}
  \caption{The results of object location recognition with the information of the object class, shape, and color using \textcolor{vg}{VG}.}
  \label{figure:object-all}
  \vspace{-1.0ex}
\end{figure}

\begin{figure}[t]
  \centering
  \includegraphics[width=0.95\columnwidth]{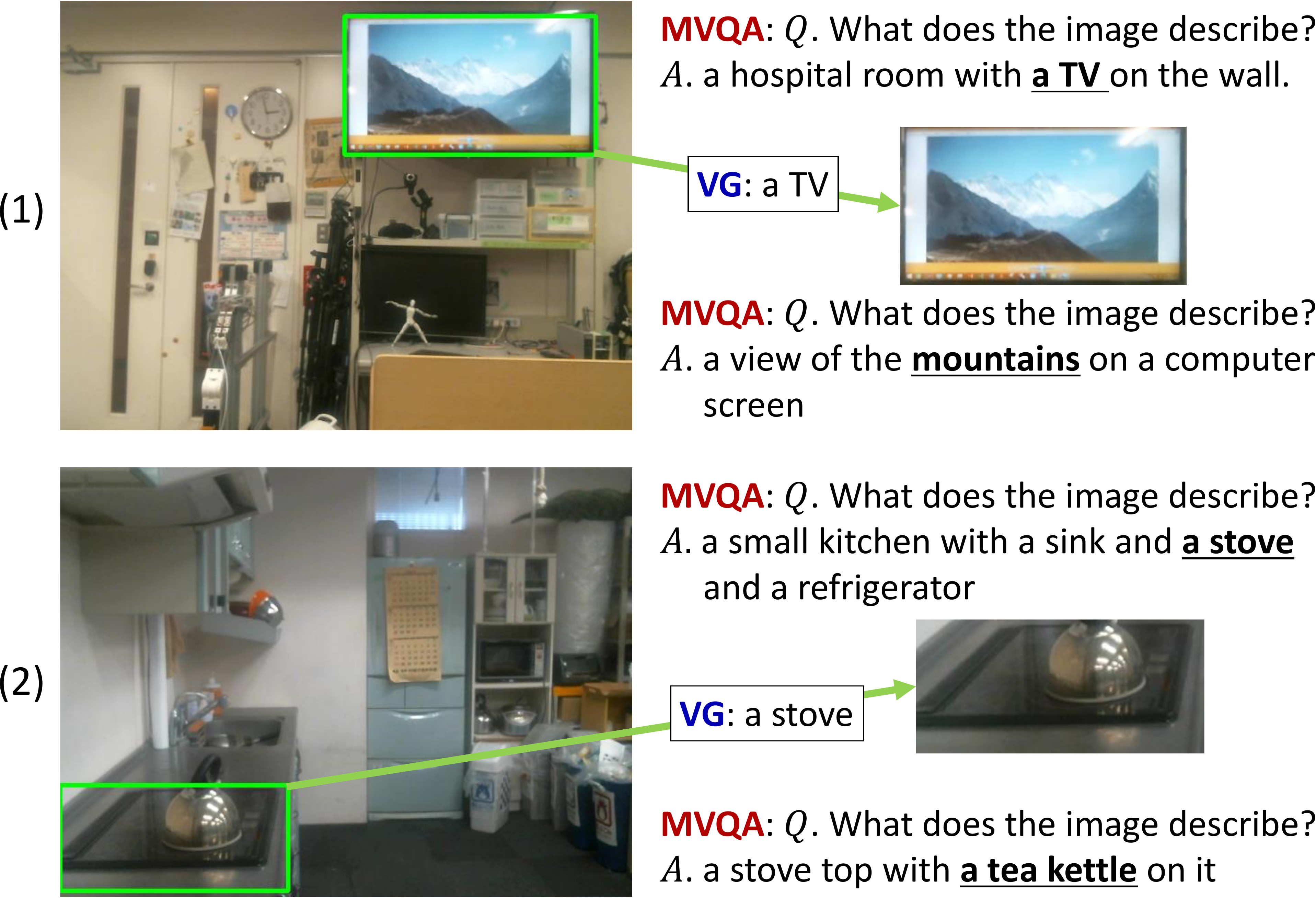}
  \vspace{-1.0ex}
  \caption{The results of stepwise refinement of object recognition using \textcolor{mvqa}{MVQA} and \textcolor{vg}{VG}, in order.}
  \label{figure:detailed}
  \vspace{-3.0ex}
\end{figure}

\subsection{Object Recognition}
\switchlanguage%
{%
  Although object recognition is partially a common problem setup, it is described for the sake of comprehensiveness.
  The essence of this study lies in the next section and thereafter.

  First, \figref{figure:object-each} shows the results of object class and feature recognition using \textcolor{mvqa}{MVQA} or \textcolor{itr}{ITR}.
  We set up problems of class recognition ``class'', shape recognition ``shape'', and color recognition ``color'', and applied them to three object images: (1) yellow round cup, (2) clear round glass, and (3) blue rectangular box.
  The questions $Q$ with four different articles in \textcolor{mvqa}{MVQA}, the choices $C$ that takes the match, and the choices $C$ in \textcolor{itr}{ITR} are prepared.
  \textcolor{mvqa}{MVQA} shows the percentage of each $C$ and Invalid (invalid answer other than the predefined $C$) for the 20 trials described in \secref{subsec:information-extraction}, and \textcolor{itr}{ITR} shows the probability of each $C$ for a single trial.
  Note that for each object, $V$ from five different angles of view are prepared, and the mean $\mu$ and standard deviation $\sigma$ are also shown for the percentage of correct responses.
  Regarding class recognition, both \textcolor{mvqa}{MVQA} and \textcolor{itr}{ITR} correctly recognize (1) as ``cup'', (2) as ``glass'', and (3) as ``box''.
  As for \textcolor{mvqa}{MVQA}, for example, various answers such as ``yellow cup'' and ``coffee cup on the desk'' are output for (1), so if a word of each $C$ is included in the answer, we judge that the output matches it.
  Next, regarding shape recognition, we prepared two $C$: ``round'' and ``rectangular''.
  (1) and (2) are correctly recognized as ``round'', and (3) as ``rectangular''.
  In \textcolor{mvqa}{MVQA}, even if there are many invalid cases, the recognition accuracy becomes 100\% by ignoring them.
  On the other hand, especially for (1), the answer of \textcolor{mvqa}{MVQA} is often invalid, and \textcolor{itr}{ITR} is able to give a clearer answer.
  Finally, for color recognition, \textcolor{mvqa}{MVQA} and \textcolor{itr}{ITR} correctly identify (1) as ``yellow'', (2) as ``clear'', and (3) as ``blue'', except for (3) in \textcolor{itr}{ITR}, almost the same as class and shape recognition.
  However, there are many invalid answers for (2) in \textcolor{mvqa}{MVQA}, and the answers for (3) in \textcolor{itr}{ITR} are highly scattered.
  In the former, the answer is often ``white'' because it has assimilated into the desk.
  In the latter, the white part of the box seems to be transparent, and the probability of ``clear'' is considered to be higher.
  Although color and shape recognition is uncommon, class recognition is a common problem.
  In \cite{matsushima2022wrs}, class recognition using \textcolor{itr}{ITR} is used in a robot competition because it is possible to improve the recognition accuracy subsequently without re-training by just tuning the text of the choices (called prompt tuning).

  Next, \figref{figure:object-all} shows the results of the object location recognition using \textcolor{vg}{VG}.
  Not only phrases for the class recognition such as ``cup'', ``glass'', and ``box'', but also phrases for the feature recognition such as ``round object'' and ``clear object'' can be used to extract the location of the corresponding object.
  As exactly the same results can be obtained from the class recognition and feature recognition phrases, the results represented by the green bounding boxes have overlapped.
  Note that since OFA outputs only one bounding box that matches best, only object (1) is recognized as ``round object''.

  Finally, \figref{figure:detailed} shows the results of stepwise refinement of object recognition.
  Two situations, (1) and (2), were prepared, and object recognition of the target object was performed in the order of \textcolor{mvqa}{MVQA}, \textcolor{vg}{VG}, and \textcolor{mvqa}{MVQA}.
  For example, in (1), the recognition is initially limited to a room with a TV on the wall, but by extracting the location of the TV and performing \textcolor{mvqa}{MVQA} on the image, it is possible to recognize that a mountain is displayed on the TV.
  More detailed object recognition is possible by stepwise refinement of the target object.
}%
{%
  本項では一般的な物体認識について述べており, 部分的には良くある問題設定であるが網羅性のため記述しており, 本研究の真髄は次項以降にある.

  まず, \figref{figure:object-each}に\textcolor{mvqa}{MVQA}または\textcolor{itr}{ITR}を用いた物体の存在と特徴の認識結果を示す.
  物体認識``object'', 形状認識``shape'', 色認識``color''という問題を設定し, それらを3つの物体画像(1) yellow round cup, (2) clear round glass, (3) blue rectangular boxに対して行った.
  \textcolor{mvqa}{MVQA}における冠詞を4種類に変化させた質問文Qと合致を取る選択肢C, \textcolor{itr}{ITR}における選択肢を記載している.
  \textcolor{mvqa}{MVQA}は\secref{subsec:information-extraction}で述べた20回の試行に関する各選択肢とInvalid(選択肢以外の無効な答え)の割合, \textcolor{itr}{ITR}は1回の試行に関する各選択肢に関する確率を示している.
  なお, それぞれの物体について5つの画角からの画像を用意し, それぞれの正解率について平均$\mu$と標準偏差$\sigma$も記載している.
  まず物体認識について, \textcolor{mvqa}{MVQA}も\textcolor{itr}{ITR}も, (1)をcup, (2)をglass, (3)をboxと, 正しく認識していることがわかる.
  \textcolor{mvqa}{MVQA}については, 例えば(1)であればyellow cupやcoffee cup on the deskなどの様々な言葉が出力されるため, 選択肢の単語が含まれていたら合致したと判断している.
  次に形状認識について, ここではroundとrectangularという2つの選択肢を用意しているが, (1)と(2)がround, (3)がrectangularと正しく認識できている.
  一方, 特に(1)については答えが無効な場合が多く, \textcolor{itr}{ITR}の方がはっきりと答えを出すことができている.
  最後に色認識について, 基本的には物体認識や形状認識と同様, (1)をyellow, (2)をclear, (3)をblueと正しく認識できている.
  ただし, (2)の\textcolor{mvqa}{MVQA}については無効な答えが多く, (3)の\textcolor{itr}{ITR}についてはその答えが大きく分散している.
  前者については机と同化して``white''と答える場合が多かった.
  また, 後者については白い部分が透過しているように見え, clearの確率が高くなったと考えられる.
  なお, 物体の色や形状を言語として抽出することは珍しいが, 物体認識自体はありふれた問題設定である.
  \cite{matsushima2022wrs}では, \textcolor{itr}{ITR}を使った物体認識はその選択肢のテキストを調整すること(prompt tuning)で認識精度を再学習無しに後から向上可能である点が良いとして, ロボット競技会に用いている.

  次に, \figref{figure:object-all}に物体の位置認識実験の結果を示す.
  物体認識で用いたcupやglass, boxだけでなく, 形や色を用いて, round objectやclear object等のフレーズでも該当する物体の位置を抽出することができる.
  緑の枠が2つから3つ重なってしまっているが, 例えばcupやround object, yellow object等からは全く同じ位置認識結果が得られる.
  なお, OFAでは最も合致する一つのバウンディングボックスしか出力されないため, round objectについては(1)の物体のみが認識されている.

  最後に, \figref{figure:detailed}に物体認識の段階的詳細化に関する実験の結果を示す.
  (1)と(2)という2つの例を用意し, それぞれ\textcolor{mvqa}{MVQA}, \textcolor{vg}{VG}, \textcolor{mvqa}{MVQA}の順で目標となる物体中の物体認識を行った.
  例えば(1)について, 始めはテレビが壁にかかった部屋であるということしか認識出来ないが, テレビ部分を位置抽出し, その画像に対してさらにVQAを行うことで, テレビ画面に山が映っているということを認識できる.
  目標となる物体の段階的詳細化による, より細かい物体認識が可能である.
}%

\begin{figure}[t]
  \centering
  \includegraphics[width=0.85\columnwidth]{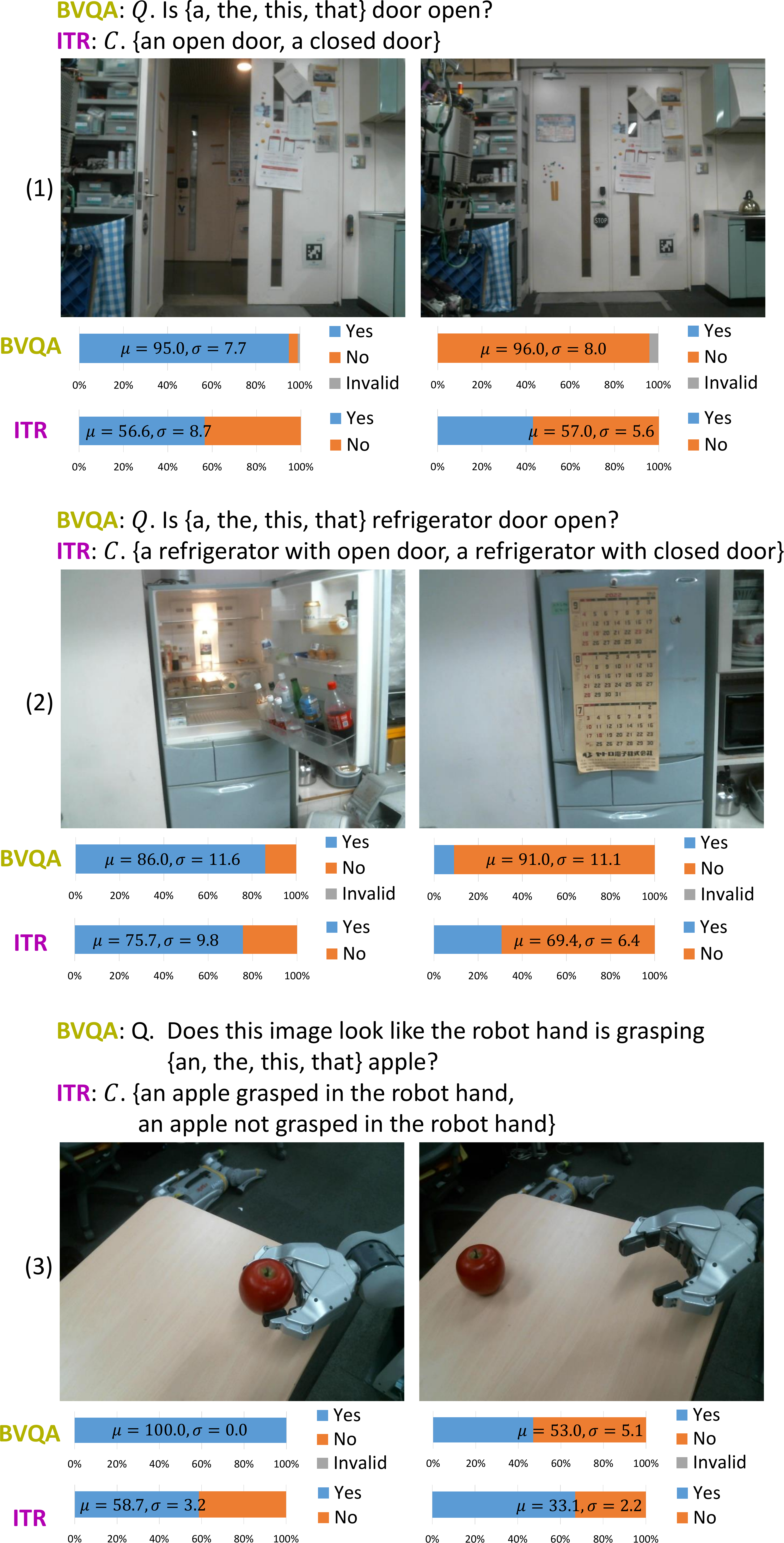}
  \vspace{-1.0ex}
  \caption{The results of state recognition using \textcolor{bvqa}{BVQA} and \textcolor{itr}{ITR} for three situations.}
  \label{figure:state}
  \vspace{-3.0ex}
\end{figure}

\begin{figure}[t]
  \centering
  \includegraphics[width=0.85\columnwidth]{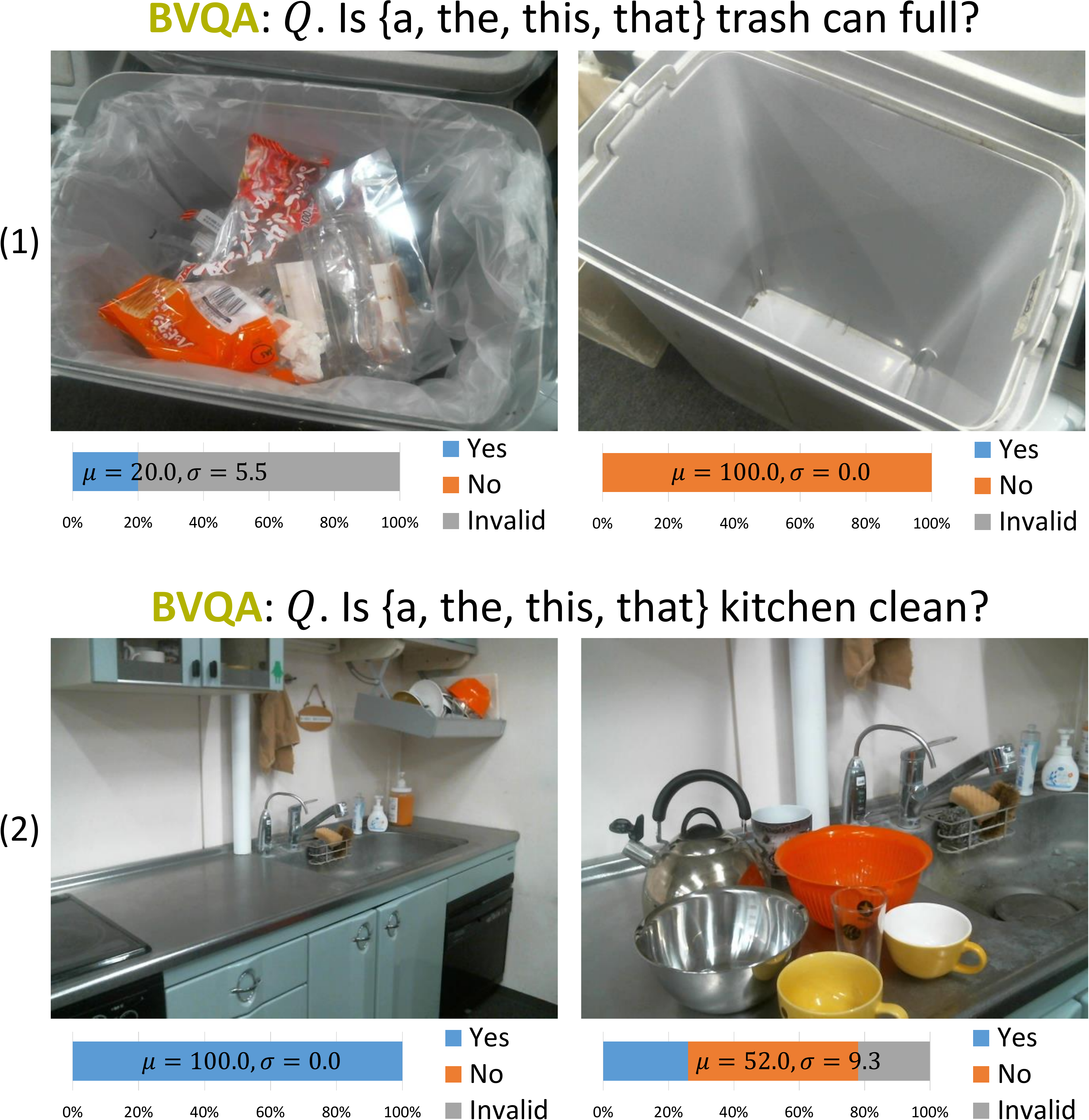}
  \vspace{-1.0ex}
  \caption{The results of qualitative state recognition using \textcolor{bvqa}{BVQA} for two situations.}
  \label{figure:quality}
  %\vspace{-3.0ex}
\end{figure}

\begin{figure}[t]
  \centering
  \includegraphics[width=0.9\columnwidth]{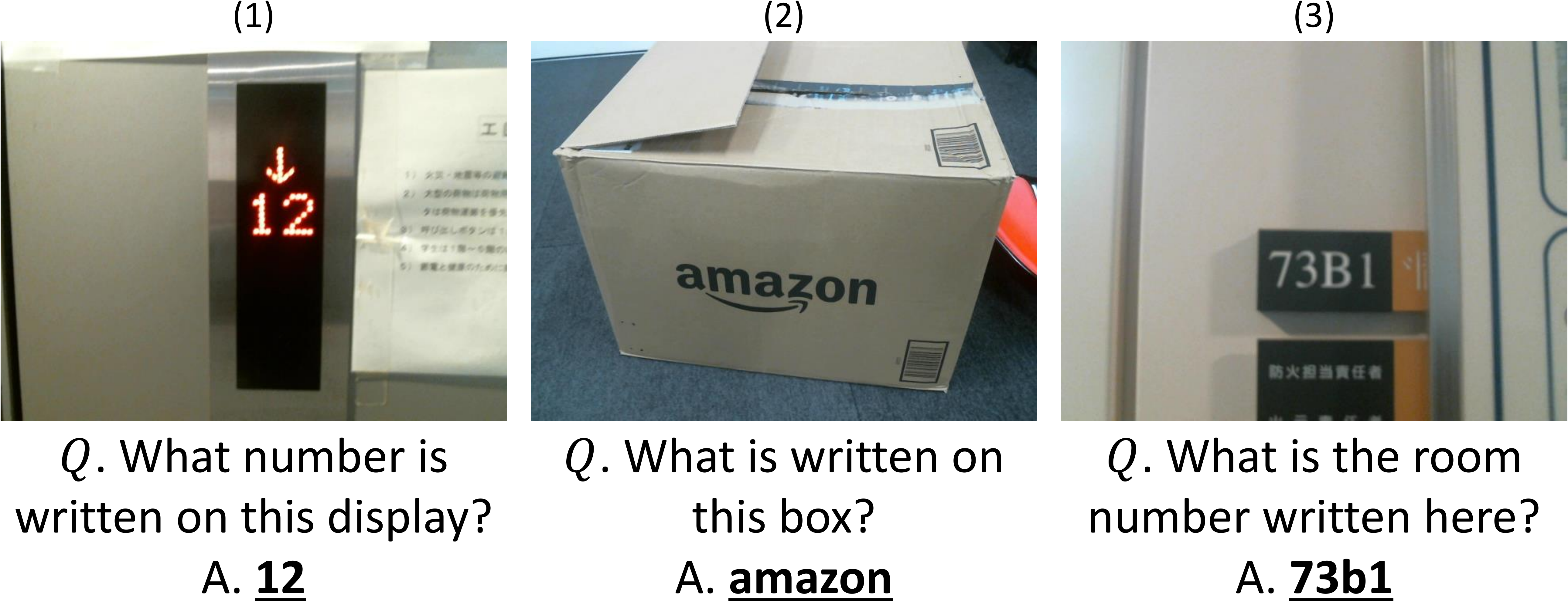}
  \vspace{-1.0ex}
  \caption{The results of character state recognition using \textcolor{mvqa}{MVQA}.}
  \label{figure:ocr}
  \vspace{-3.0ex}
\end{figure}

\subsection{State Recognition}
\switchlanguage%
{%
  \figref{figure:state} shows the results of binary state recognition using \textcolor{bvqa}{BVQA} or \textcolor{itr}{ITR}.
  Regarding \textcolor{bvqa}{BVQA}, we describe four types of $Q$ with different articles, and regarding \textcolor{itr}{ITR}, we describe two $C$.
  \textcolor{bvqa}{BVQA} shows the percentages of Yes, No, and Invalid (answers other than Yes or No) for the 20 trials described in \secref{subsec:information-extraction}, and \textcolor{itr}{ITR} shows the probabilities of Yes (the first choice) and No (the second choice) for one trial.
  Note that for each state, $V$ from five different angles of view are prepared, and the mean $\mu$ and standard deviation $\sigma$ are also shown for the percentage of correct responses.
  Three examples (1)--(3) are provided, where all the answers should be Yes for the left image and No for the right image.
  From the figures, \textcolor{bvqa}{BVQA} correctly outputs Yes or No in all cases.
  On the other hand, the ratio of Yes and No for \textcolor{itr}{ITR} is close to 50\% and 50\%, and it is not as clearly recognized as \textcolor{bvqa}{BVQA}.
  As for (3), the answer is reversed.
  \textcolor{bvqa}{BVQA} is easier to use to describe the state of the environment and objects than \textcolor{itr}{ITR}, because \textcolor{bvqa}{BVQA} allows more flexible question phrases and clearer answers than \textcolor{itr}{ITR}.
  Note that in (1) and (2), $Q$ is asked in the form of ``Is -?'' but by asking $Q$ of ``Does this image look like -?'' as in (3), it is possible to make the percentage of invalid answers almost zero.
  For (3), if we ask $Q$ of ``Is -?'', 20/20 cases are invalid.
  On the other hand, it has been experimentally found that the answers to $Q$ of ``Does this image look like -?'' is not very accurate compared to ``Is -?''.

  Of course, it is possible to build state recognizers with almost 100\% accuracy by collecting training data or by manually programming for the specific environment.
  However, an important contribution of this study is that it can be easily applied to a wide variety of environments and robots, without any programming or model training specific to each environment and robot.

  Next, \figref{figure:quality} shows the results of binary state recognition using \textcolor{bvqa}{BVQA} for qualitative states that are more difficult to judge.
  The percentages of Yes, No, and Invalid are shown for $Q$ of (1) whether the trash can is full and (2) whether the kitchen is clean.
  Although the results are not as clear as those of \figref{figure:state}, answers are generated in a manner consistent with human intuition.
  It can be applied to tasks such as robot patrolling.

  Finally, \figref{figure:ocr} shows the results of character state recognition using \textcolor{mvqa}{MVQA}.
  The robot can recognize various character shapes such as numbers on an elevator display, letters on a cardboard box, room numbers, and so on, by devising questions.
  This can be applied to robots checking mail, recognizing the floor number in an elevator, and entering a certain room based on the room number.
}%
{%
  \figref{figure:state}に, \textcolor{bvqa}{BVQA}または\textcolor{itr}{ITR}による二値状態認識の結果を示す.
  \textcolor{bvqa}{BVQA}については冠詞を4種類に変化させた質問文, \textcolor{itr}{ITR}については2つの選択肢を提示している.
  \textcolor{bvqa}{BVQA}は\secref{subsec:information-extraction}で述べた20回の試行に関するYes, No, Invalid(Yes/No以外の無効な答え)の割合, \textcolor{itr}{ITR}は1回の試行に関するYes(選択肢1つ目)とNo(選択肢2つ目)に関する確率を示している.
  なお, それぞれの状態について5つの画角からの画像を用意し, それぞれの正解率について平均$\mu$と標準偏差$\sigma$も記載している.
  (1)--(3)の3つの例を用意しているが, 左図は全て答えがYesであるべき, 右図は全て答えがNoであるべき画像となっている.
  図から, 全ての例について\textcolor{bvqa}{BVQA}は正しくYesまたはNoを出力している.
  一方で, \textcolor{itr}{ITR}については全体的にYesとNoの比率が50\%に近く, \textcolor{bvqa}{BVQA}ほど明確に認識できているわけではない.
  (3)については答えが逆転してしまっている.
  このように, 環境や物体の状態を表現することは, \textcolor{itr}{ITR}に比べて\textcolor{bvqa}{BVQA}の方が柔軟な質問文が可能で答えも明確であり使いやすい.
  なお, (1)と(2)では質問を``Is -?''という形で行ったが, (3)のように``Does this image look like -?''という質問をすることで, invalidの割合をほぼ0とすることもできる.
  (3)については, ``Is -?''で質問すると20/20がinvalidとなってしまった.
  一方で, ``Does this image look like -?''の形は認識精度があまり高くないことも実験的に分かっており, 本研究では基本的に``Is -?''という質問方法を取っている.

  もちろん本実験のような状態認識は, その環境において学習データを集めたり, 人間が手動で認識プログラムを組んだりすることで, 認識率ほぼ100\%の認識器を組むこともできる.
  しかし本研究で重要なのは, 汎用性や適応性を損なう, その環境やロボットに合ったプログラム記述やモデル学習を一切行わない点であり, 容易に多様な環境やロボットに適用できる点である.

  次に, より判断の難しい質的な状態について, \textcolor{bvqa}{BVQA}を用いて二値状態分類を行った結果を\figref{figure:quality}に示す.
  (1)ゴミ箱がいっぱいか, (2)キッチンは綺麗かという質問に対して, Yes, No, Invalidの割合を表示している.
  \figref{figure:state}ほど明確ではないが, 人間の直感と一致した形でYesまたはNoが生成されており, ロボットの見回りタスク等に応用できる.

  最後に, 文字状態認識に関する実験を\figref{figure:ocr}に示す.
  エレベータのディスプレイに映った数字, ダンボールに書かれた文字, 部屋番号など, 様々な形状をした文字の認識を, 質問文を工夫することで行うことができる.
  ロボットが郵便物を確認したり, エレベータの階を認識して行動したり, 指令された部屋に確認してから入るといったことに応用できる.
}%

\begin{figure}[t]
  \centering
  \includegraphics[width=0.9\columnwidth]{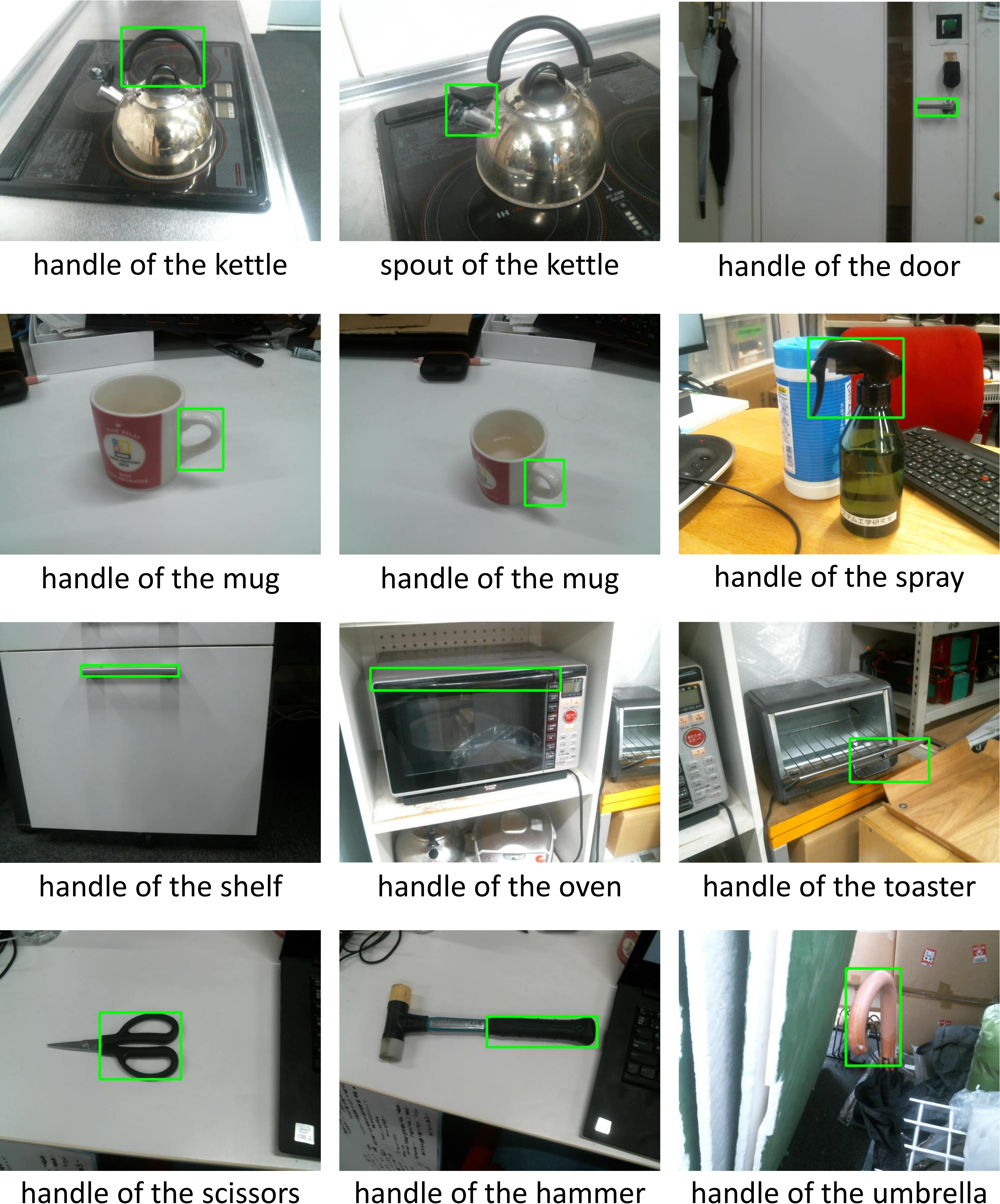}
  \vspace{-1.0ex}
  \caption{The results of affordance recognition using \textcolor{vg}{VG}.}
  \label{figure:affordance}
  \vspace{-1.0ex}
\end{figure}

\subsection{Affordance Recognition}
\switchlanguage%
{%
   \figref{figure:affordance} shows the results of the affordance recognition using \textcolor{vg}{VG}.
  First, the handle and the spout of the kettle are recognized in the top row of \figref{figure:affordance}.
  This means that the robot can grasp the kettle, bring the spout close to the cup, and pour hot water into the cup.
  In addition, the robot can recognize the grasping positions of various objects such as a door, cup, spray, shelf, oven, toaster, scissors, hammer, and umbrella.
}%
{%
  \figref{figure:affordance}にアフォーダンス認識実験の結果を示す.
  まず最上段について, やかんのハンドルと注ぎ口が認識できている.
  これにより, やかんを把持して, コップまで注ぎ口を近づけ, お湯を注ぐことが可能になる.
  この他にも, ドアやコップ, スプレー, 棚, オーブン, トースター, はさみ, ハンマー, 傘など, 多様な物体の把持位置を認識することが可能であった.
}%

\begin{figure}[t]
  \centering
  \includegraphics[width=0.9\columnwidth]{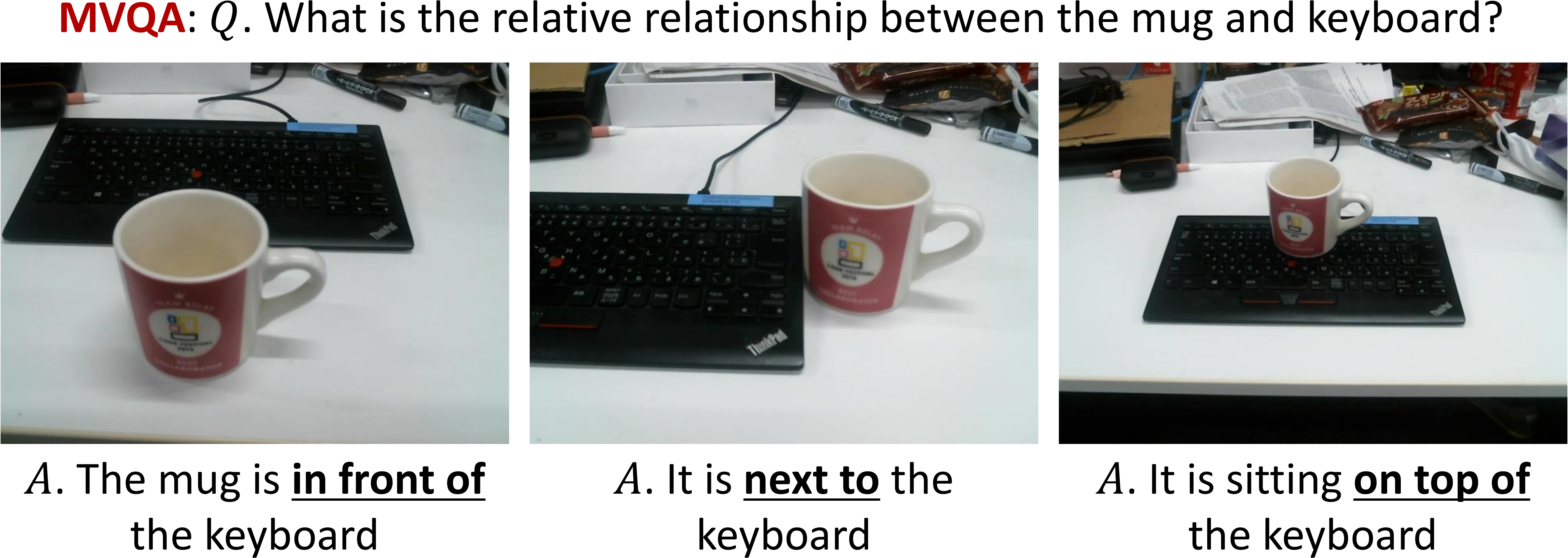}
  \vspace{-1.0ex}
  \caption{The results of relation recognition using \textcolor{mvqa}{MVQA} for three situations.}
  \label{figure:relation}
  \vspace{-3.0ex}
\end{figure}

\subsection{Relation Recognition}
\switchlanguage%
{%
  \figref{figure:relation} shows the results of relation recognition using \textcolor{mvqa}{MVQA}.
  We prepared $V$ where the keyboard and mug are in front-back, left-right, and top-bottom positional relationships, and asked $Q$ of ``What is the relative relationship between the mug and the keyboard?''.
  Each relationship is expressed by the prepositions ``in front of'', ``next to'', or ``on top of''.
  By matching these prepositions with predefined $C$, we can recognize the relationship and apply it to robotic manipulation.
  % In addition, the change in this relationship can be used to automatically segment a long behavior.
}%
{%
  \figref{figure:relation}に関係認識実験の結果を示す.
  keyboardとmugが前後, 左右, 上下という状態の画像を用意し, ``what is the relative relationship between the mug and keyboard?''という質問を行った.
  それぞれの関係性が, ``in front of'', ``next to'', ``on top of''という前置詞によって表現されている.
  これらを選択肢からマッチングすることで, その関係を認識し, マニピュレーションに応用可能である.
  % また, この関係性の変化から動作の分節化を行うことができる.
}%

\begin{figure}[t]
  \centering
  \includegraphics[width=0.9\columnwidth]{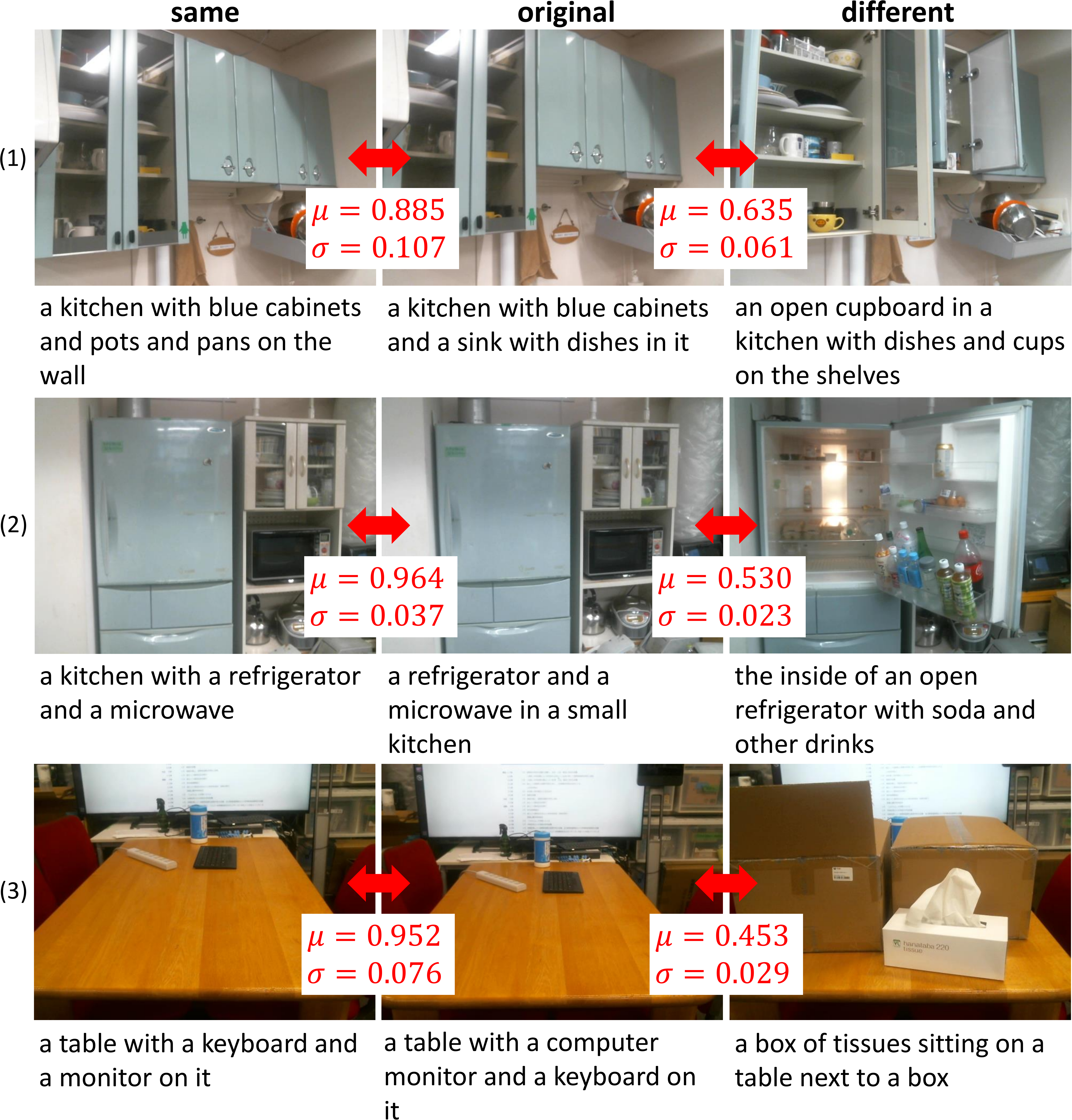}
  \vspace{-1.0ex}
  \caption{The results of anomaly detection using \textcolor{dic}{DIC} for three situations. The original image and the images that are ``same'' as or ``different'' from it are shown, along with the similarity scores.}
  \label{figure:anomaly}
  \vspace{-3.0ex}
\end{figure}

\subsection{Anomaly Detection}
\switchlanguage%
{%
  \figref{figure:anomaly} shows the results of anomaly detection using \textcolor{dic}{DIC}.
  We take ``original'' as an original $V$, and prepare a set of $V$ that are almost the same (``same'') or different (``different'') with the original $V$.
  We show the description of those images output by IC and the cosine similarity between ``original'' and ``same'', or ``original'' and ``different''.
  Note that five images applied with RGBShift are prepared for each $V$, and the average $\mu$ and standard deviation $\sigma$ of the similarity are described.
  Three examples (1)--(3) are given, all of which show high similarity between ``original'' and ``same'', and low similarity between ``original'' and ``different''.
  Opening the door of the shelf or placing objects on the desk increases the number of descriptions about the objects in the shelf or the newly placed objects.
  Although the results output by IC change slightly when the angle of view and lighting conditions change, the similarity does not decrease much because the descriptions are similar.
  This similarity enables the robot to detect anomalies and to communicate them to the surrounding people through spoken language.
}%
{%
  \figref{figure:anomaly}に異常検知実験の結果を示す.
  ``original''を元画像として, その画像とほとんど同じ状態(``same''), または異なる状態(``different'')を用意する.
  それら画像のICにおける記述, ``original''と``same'', または``original''と``different''の間のコサイン類似度を示している.
  なお, それぞれの画像に対して, RGBShiftを適用した5つの画像を用意し, 類似度の平均$\mu$と標準偏差$\sigma$を記述している.
  (1)--(3)の3つの例を挙げるが, どれも``original''と``same''の間の類似度は高く, ``original''と``different''の間の類似度はそれに比べると低い.
  棚のドアを開けたり机の上に物を置いたりすることで, 棚の中の物体や新しく設置された物体に関する記述が増える.
  ICの結果は画角や照明条件が変わると多少変化するが, 記述は似たものであるため, 類似度はあまり下がらない.
  この類似度から異常検知が可能であり, その異常を言語によって周囲に伝えることができる.
}%

\section{Discussion} \label{sec:discussion}
\switchlanguage%
{%
  The experimental results are summarized and discussed in terms of their performance, limitations, and development.
  First, object recognition was possible using either \textcolor{mvqa}{MVQA} or \textcolor{itr}{ITR}. %, and \textcolor{itr}{ITR} was able to output clearer answers compared to \textcolor{mvqa}{MVQA}.
  In \textcolor{mvqa}{MVQA}, there are some cases where the answer becomes invalid, but this can be improved by preparing multiple questions and images, ignoring the invalid answers, and obtaining the answer with the highest probability.
  While shape and color recognition of objects can be performed in the same way, there were some cases where recognition does not work well with \textcolor{itr}{ITR}.
  It was also possible to recognize the location of an object not only by its class, but also by its shape and color.
  Furthermore, the combination of \textcolor{vg}{VG} and \textcolor{mvqa}{MVQA} enables stepwise refinement of object recognition.

  Second, state recognition was possible using either \textcolor{bvqa}{BVQA} or \textcolor{itr}{ITR}, and \textcolor{bvqa}{BVQA} was able to output clearer answers compared to \textcolor{itr}{ITR}.
  In \textcolor{itr}{ITR}, information regarding the object (e.g. door) is strong, while the information regarding the state (e.g. open) is weak, making it difficult to judge the state.
  As for \textcolor{bvqa}{BVQA}, there may be many invalid answers depending on the question, but this can be avoided by adjusting the question.
  This state recognition can also answer to qualitative questions such as whether the kitchen is clean or dirty, which is less clear compared to whether the door is open or closed.
  Of course, this is a concept averaged from a large dataset, but it is possible that the robot can acquire a general human sense.
  In addition, the robot can recognize the character state on a display or on paper, and these can greatly expand the range of executable tasks without any re-training.

  Third, affordance recognition was possible by the textual innovations in \textcolor{vg}{VG}.
  The concept of affordance is obtained implicitly from a large dataset without using a specialized neural network.
  Affordance recognition is possible for a wide variety of objects and tools.

  Fourth, relation recognition was possible by taking prepositional matches with predefined choices using \textcolor{mvqa}{MVQA}.
  By knowing the relation between specified objects, the robot can manipulate them to change to the target relation.
  In the future, it is expected to be applied to motion segmentation, human motion understanding, imitation learning, and so on, based on the change of relations between objects and bodies.

  Finally, anomaly detection was possible by vectorization and differencing of sentences output by IC (\textcolor{dic}{DIC}).
  Although the outputs of IC change slightly with changes in the angle of view or with noise, the differences in the descriptions are not large, resulting in a high degree of similarity in the outputs.
  On the other hand, when the images change significantly, the outputs of IC also change significantly and the similarity score drops, thus enabling anomaly detection with the threshold value.
  Using sentences to detect differences is compatible with the use of chat tools and conversation to report anomalies, and further applications are expected.

  In this study, we have comprehensively described the innovativeness of this idea through preliminary experiments.
  The vision-language model is still in the process of significant development, and it is necessary to keep an eye on its trend, create further applications, and incorporate them into actual robots.
  As one example, we show a simple experiment in a supplementary video.
  We have performed an automatic patrol experiment using state recognition on the mobile robot Fetch.
  The robot can close the refrigerator door by \textcolor{itr}{ITR} if it is open, turn off the faucet by \textcolor{bvqa}{BVQA} if water is running, and exit the room by \textcolor{itr}{ITR} if the door is open.
  On the other hand, more detailed performance checks are needed for each example when incorporating \textbf{PTVLMs} into actual robots, which will be addressed in the future.
  In particular, it is important to discuss how to prepare the question phrases, predefined choices, and thresholds for determining the output.
  Although we have adjusted them manually in the preliminary experiments, it would be better to automatically acquire them from the data of actual tasks.
  \cite{kawaharazuka2023ofaga} has proposed a method to create multiple questions by changing the articles, state expressions, forms, and wordings, and to select the best combination of questions by a genetic algorithm.
  In the future, we would like to generalize the process of choosing questions and predefined choices for the best performance.
  % It is important to use the pre-trained model without re-training the entire network so as not to reduce the versatility and adaptability of the recognition.
  % We believe that this will facilitate the improvement of recognition abilities for various robots and environments, and will revolutionize the future of robotics.
}%
{%
  実験結果についてその性能と限界, 発展をまとめ考察する.
  まず, 物体認識は\textcolor{mvqa}{MVQA}または\textcolor{itr}{ITR}により可能であったが, \textcolor{itr}{ITR}の方がより明確に答えを出力することが可能であった.
  \textcolor{mvqa}{MVQA}を用いた場合は意図していない答えが出力され回答が無効になるケースがある程度存在するが, 質問と画像を変化させて割合を取ることで対応することが可能である.
  物体の形状認識や色認識も同様に行うことが可能であった.
  一方で, \textcolor{itr}{ITR}では一部認識が上手く行かないケースも見受けられた.
  物体の位置認識は物体の名称だけでなく形状や色によっても可能であった.
  さらに, \textcolor{vg}{VG}と\textcolor{mvqa}{MVQA}を組み合わせることで物体認識の段階的詳細化が可能である.

  次に, 状態認識は\textcolor{bvqa}{BVQA}または\textcolor{itr}{ITR}により可能であったが, \textcolor{bvqa}{BVQA}の方がより明確に答えを出力することが可能であった.
  \textcolor{itr}{ITR}における合致するテキストの検索では, 物体(例えばドア)に関する情報が強く, 状態(開いている)に関する情報が薄まってしまい, 状態の判断が難しい.
  \textcolor{bvqa}{BVQA}については, 質問文によっては無効な答えが多い場合もあるが, 質問文の調整によりこれを回避することができる.
  そしてこの状態認識は, ドアが開いているか閉まっているかほど明確ではない, キッチンが綺麗か汚いかといった質的な質問にも答えることができる.
  もちろんこれは大規模なデータセットから得られた平均化された概念ではあるが, 一般的な人間の感覚を得ることができている可能性が高い.
  加えて, ディスプレイや紙に書かれた文字状態も認識することができ, 一切の再学習無しにタスクの幅を大きく広げることができると考えられる.
  % 今後ロボットプログラミングのif-elseやassertが全て言語で記述される可能性も考えられる.

  次に, アフォーダンス認識は\textcolor{vg}{VG}におけるテキストの工夫によって可能であった.
  これまでの専用ネットワークを用いず, 大規模なデータセットから暗黙的にアフォーダンスの概念が獲得されている.
  多様な物体についてアフォーダンス認識が可能であり, ロボットのマニピュレーションを大きく革新できる可能性が高い.

  次に, 関係認識は\textcolor{mvqa}{MVQA}により前置詞の合致を取ることで可能であった.
  指定した物体間の関係を知ることで, それらを目標状態に変化させるマニピュレーションが可能である.
  また, 今後物体間関係変化に基づいた分節化や動作理解, 模倣学習等への応用が期待される.

  最後に, 異常検知はIC出力のベクトル化と差分(\textcolor{dic}{DIC})により可能であった.
  ICの結果は画角が変化することで多少変わるが, 文章としての差は大きくないため高い類似度が出力される.
  一方で, 大きく画面が変わるとICの結果も大きく変化し, 類似度が下がるため閾値から異常検知が可能となる.
  文章で差を取ることで, チャットや会話を使った異常の報告等とも相性が良く, 今後さらなる発展が見込める.

  本研究では予備的な実験により網羅的にこの新しい考え方の革新性について述べた.
  一方で, 視覚-言語モデルは現在大きく発展している途中であり, その動向を見極め, さらなる応用事例を創出し, 実際のロボットに取り込んでいく必要がある.
  その一例として, ....
  また, 実際のロボットに組み込む際に, 一つ一つの例に対してより綿密な性能の確認が必要であり, 今後の課題とする.
  加えて, \textcolor{mvqa}{MVQA}/\textcolor{bvqa}{BVQA}における質問文やその出力を決定する閾値, \textcolor{itr}{ITR}の閾値等は, 今後実際のタスク中のデータから学習してい必要があると考える.
  一方で, 重要なのは汎用性や適応性を下げないために, ネットワーク全体の再学習はせず, あくまで事前学習モデルを利用する点であると考える.
  これは, 様々なロボットや環境における容易な認識能力の向上を促し, 今後のロボットを大きく変革することができると考える.
}%

\section{CONCLUSION} \label{sec:conclusion}
\switchlanguage%
{%
  In this study, we comprehensively and experimentally show that a pre-trained vision-language model can be used for various recognition behaviors in robots.
  The model is used to extract discrete values of a few choices or continuous values of a few dimensions, without directly connecting them to the action of robots.
  We classify the information extraction into five methods: binary state recognition using VQA (\textcolor{bvqa}{BVQA}), matching with predefined textual choices after VQA (\textcolor{mvqa}{MVQA}), retrieval of matching textual choices from images (\textcolor{itr}{ITR}), extraction of locations in images that match the text (\textcolor{vg}{VG}), and difference computation of outputs by IC (\textcolor{dic}{DIC}).
  Using these five methods, robots are able to perform object recognition including class, shape, and color, as well as location recognition of objects, state recognition including qualitative states, affordance recognition of objects and tools, relation recognition of objects, and anomaly detection.
  Although the idea is very simple, the appropriate combination of vision and language can improve the flexibility and usability of recognition systems that have been trained or programmed for specific robot bodies and environments.
  The robot can figure out whether the door is open or closed and where to grasp the tool using spoken language, and detect anomalies in a descriptive manner.
  % We hope that this study will contribute to the development of robotics for daily life support and nursing care support.
}%
{%
  本研究では, 事前学習済み視覚-言語モデルを用いてロボットの多様な認識行動が可能であることを網羅的かつ実験的に示した.
  学習モデルはロボットの行動とは直接結び付けず, 少数選択肢の離散値または少数次元の連続値を抽出するために用いる.
  この情報抽出をVQAを用いた二値判断(\textcolor{bvqa}{BVQA}), VQA後のテキスト選択肢とのマッチング(\textcolor{mvqa}{MVQA}), 画像からの合致するテキスト選択肢の検索(\textcolor{itr}{ITR}), テキストに合致する画像上の位置抽出(\textcolor{vg}{VG}), ICの差分計算(\textcolor{dic}{DIC})の5つに分類した.
  これら5つを用いて, 形や色認識, 段階的な詳細化を含んだ物体認識, 質的な値や文字まで含んだ状態認識, 物体のアフォーダンス認識, 物体の関係認識, 異常検知が可能であった.
  その考え方は非常にシンプルであるが, これまでロボットや環境ごとに学習・記述していた認識器を, 画像と言語を適切に組み合わせることで大きく革新することができる.
  ドアの開閉や電気のオンオフは言語で質問し, 道具や物体を把持する位置も言語で場所を問い, 異常検知は言語で説明的に行うことができる.
  本論文がロボットの日常生活支援や介護支援等の発展に寄与すれば幸いである.
}%

{
  %\footnotesize
  %\small
  %\bibliographystyle{junsrt}
  \bibliographystyle{IEEEtran}
  \bibliography{main}
}

\end{document}